\documentclass{article}

\usepackage[final]{neurips_2021}

\usepackage[utf8]{inputenc} %
\usepackage[T1]{fontenc}    %
\usepackage{hyperref}       %
\usepackage{url}    %
\usepackage{natbib}
\setcitestyle{square,numbers}
\usepackage{booktabs}       %
\usepackage{amsfonts}       %
\usepackage{nicefrac}       %
\usepackage{microtype}      %
\usepackage{lipsum}
\usepackage{xcolor}         %
\usepackage{todonotes}
\usepackage{sidecap}
\usepackage[export]{adjustbox}
\usepackage{wrapfig}

\usepackage{algorithm}
\usepackage{algorithmic}
\usepackage{enumitem}
\setitemize{noitemsep,topsep=0pt,parsep=0pt,partopsep=0pt,leftmargin=12pt}

\usepackage{amsmath, amssymb, amsthm,bm}
\usepackage{mathtools}
\usepackage{cancel}

\newcommand{\ve}{\mathbf e}
\newcommand{\vv}{\mathbf v}
\newcommand{\vu}{\mathbf u}
\newcommand{\vx}{\mathbf x}
\newcommand{\vz}{\mathbf z}
\newcommand{\vf}{\mathbf f}
\newcommand{\vg}{\mathbf g}
\newcommand{\gd}{|\mathbf g|/d}
\newcommand{\half}{\nicefrac{1}{2}}
\newcommand{\eps}{\epsilon}
\newcommand{\epshalf}{\nicefrac{\eps}{2}}
\newcommand{\highlight}[1]{%
  \colorbox{red!20}{$\displaystyle#1$}}

\title{Hamiltonian Dynamics with Non-Newtonian Momentum for Rapid Sampling}

\author{
  Greg {Ver Steeg}
  \qquad \qquad \qquad
  Aram Galstyan \\
  University of Southern California, Information Sciences Institute \\
  \texttt{\{gregv,galstyan\}@isi.edu}
}

\begin{document}
\maketitle

\begin{abstract}
Sampling from an unnormalized probability distribution is a fundamental problem in machine learning with applications including Bayesian modeling, latent factor inference, and energy-based model training. After decades of research, variations of MCMC remain the default approach to sampling despite slow convergence. Auxiliary neural models can learn to speed up MCMC, but the overhead for training the extra model can be prohibitive. 
We propose a fundamentally different approach to this problem via a new Hamiltonian dynamics with a non-Newtonian momentum. 
In contrast to MCMC approaches like Hamiltonian Monte Carlo, no stochastic step is required. 
Instead, the proposed deterministic dynamics in an extended state space exactly sample the target distribution, specified by an energy function, under an assumption of ergodicity. 
Alternatively, the dynamics can be interpreted as a normalizing flow that samples a specified energy model without training. 
The proposed Energy Sampling Hamiltonian (ESH) dynamics have a simple form that can be solved with existing ODE solvers, but we derive a specialized solver that exhibits much better performance.
ESH dynamics converge faster than their MCMC competitors enabling faster, more stable training of neural network energy models. 
\end{abstract}

\begin{wrapfigure}{r}{0.5\textwidth}
  \begin{center}
  \vspace{-15mm}
    \includegraphics[width=0.48\textwidth]{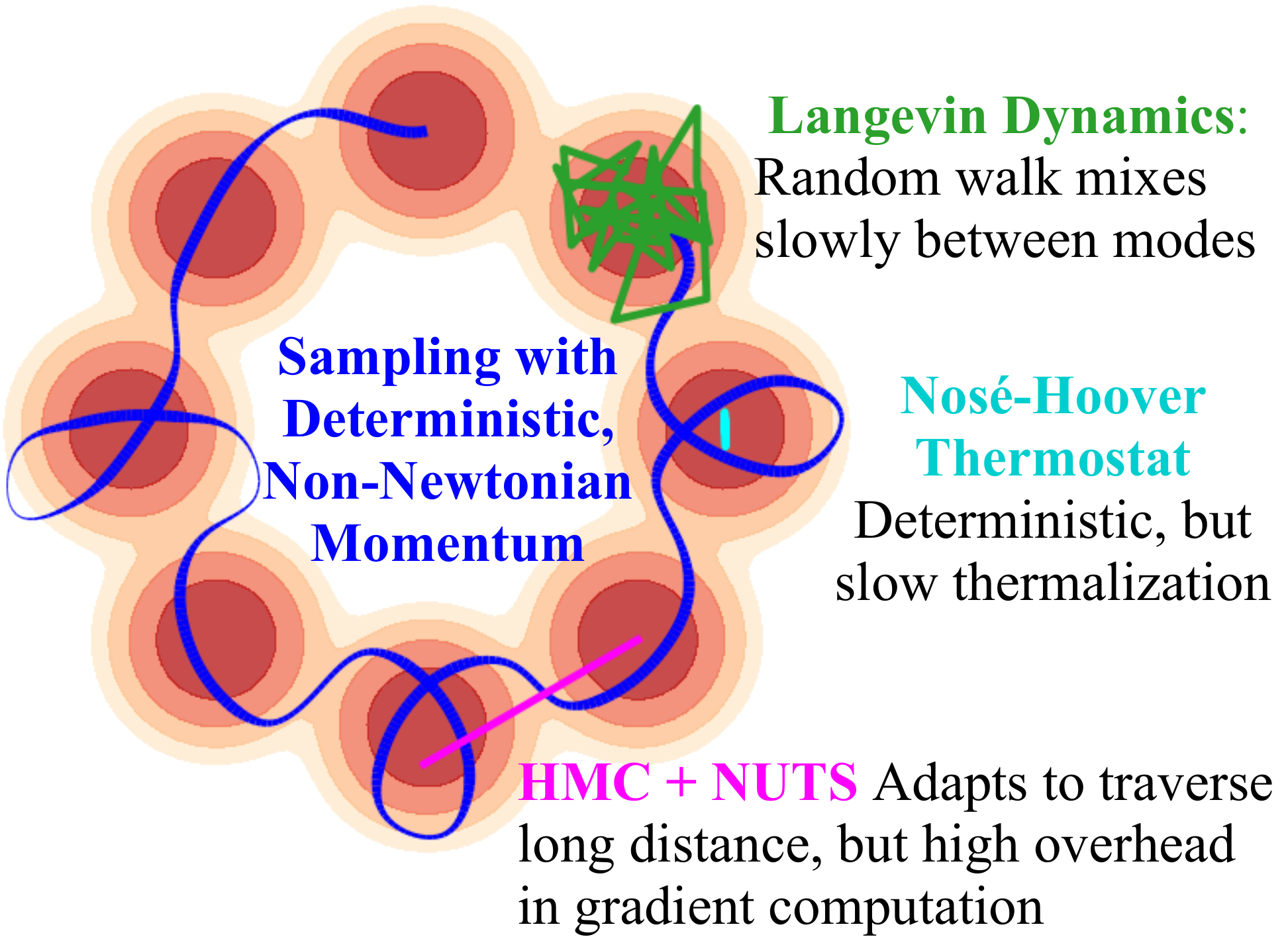}
  \end{center}
  \caption{Sampling with 50 gradient evaluations per method. A special property of our proposed deterministic dynamics using a non-Newtonian momentum, proportional to line thickness, is that it ergodically samples from the target distribution.}    \label{fig:abstract}
\end{wrapfigure}

\section{Introduction} 
While probabilistic reasoning is crucial for science and cognition~\cite{jaynes}, distributions that can be directly sampled are rare. Without sampling it is difficult to measure likely outcomes, especially in high dimensional spaces. A general purpose method to sample any target distribution, the Metropolis algorithm for Markov Chain Monte Carlo (MCMC), is considered one of the top algorithms of the 20th century~\cite{dongarra2000guest}. The ``Monte Carlo'' in MCMC refers to the essential role of stochasticity, or chance, in the approach: start with any state, propose a random update, and accept it according to a weighted coin flip. 

The most widely used methods for sampling mainly differ in their choice of proposals for the next state of the Markov chain. The most prominent example, Hamiltonian Monte Carlo (HMC), uses Hamiltonian dynamics to suggest proposals that quickly move over large distances in the target space~\cite{neal2011mcmc}. HMC using the No U-Turn Sampling (NUTS) heuristic for choosing hyper-parameters~\cite{nuts} forms the backbone of modern probabilistic programming languages like Stan~\cite{stan}. In the limit of one discrete HMC step per proposal, we recover the widely used discrete Langevin dynamics~\cite{neal2011mcmc,sgld}. 
Many approaches train neural networks to improve sampling either through better proposal distributions or flexible variational distributions~\cite{neutra,levy2017generalizing,coopnet,proposal_entropy,song2018learning,salimans2015markov,grathwohl2020no,kumar2019maximum}, but these can be expensive if, for instance, the target distribution to be sampled is itself being optimized.  

We propose a new Hamiltonian dynamics, based on the introduction of a non-Newtonian momentum, which leads to a deterministic dynamics that directly samples from a target energy function.
The idea that deterministic dynamics in an extended space can directly sample from a Gibbs-Boltzmann distribution goes back to Nos\'e~\cite{nose1984molecular}.
This principle is widely used in molecular dynamic (MD) simulations~\cite{leimkuhler2016molecular} but has had limited impact in machine learning.
One reason is that direct application of techniques developed for MD often exhibit slow mixing in machine learning applications.

Our contribution stems from the recognition that many of the design considerations for molecular dynamics samplers are to enforce physical constraints that are irrelevant for machine learning problems.
By returning to first principles and discarding those constraints, we discover a Hamiltonian whose dynamics \emph{rapidly} sample from a target energy function by using a non-physical form for the momentum.
This is in contrast to the more physical, but slower, alternative of introducing auxiliary ``thermostat'' variables that act as a thermal bath~\cite{hoover1985canonical,yoshida1990construction,martyna1992nose,martyna1996explicit}.
The result is a fast and deterministic drop-in replacement for the MCMC sampling methods used in a wide range of machine learning applications. Alternatively, our method can be viewed as a normalizing flow in an extended state space which directly samples from an unnormalized target density without additional training. 

The most significant impact of our approach is for applications where minimizing memory and computation is the priority. 
For training energy-based models, for example, sampling appears in the inner training loop. Re-training a neural sampler after each model update is costly and entails complex design and hyper-parameter choices. While MCMC sidesteps the training of an extra model, MCMC converges slowly and has led to widespread use of a number of dubious heuristics to reduce computation~\cite{nijkamp1,nijkamp2}. Because our energy sampling dynamics does not have the stochastic, random walk component of MCMC, it converges towards low energy states much faster, especially in the transient regime, and explores modes faster as shown in Fig.~\ref{fig:abstract} and demonstrated in experiments.

\section{Energy Sampling Hamiltonian (ESH) Dynamics}
\label{sec:approach}
\begin{figure}[b]
    \centering
    \includegraphics[width=0.95\textwidth,trim={0cm 1cm 0cm 10.4cm},clip]{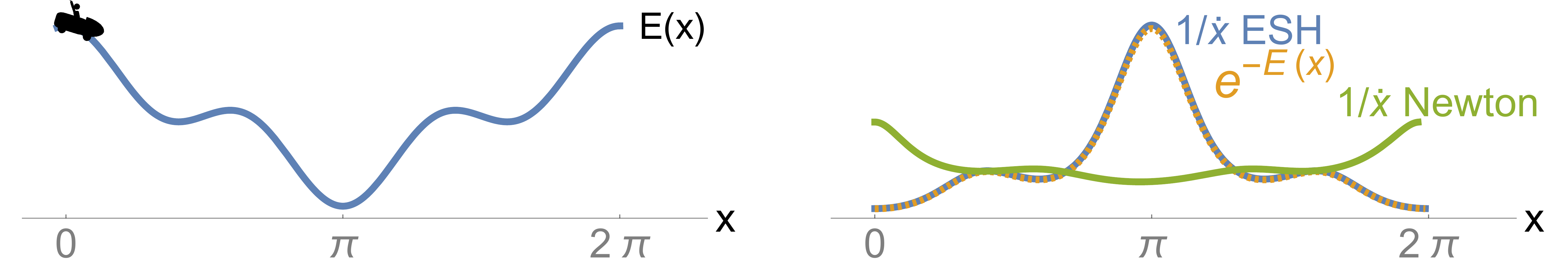}
    \caption{For Newtonian dynamics, the roller coaster spends most of its time in high energy states (left). For ESH dynamics, the time spent in the region $(x, x + dx)$ is $1/\dot x$ which is exactly proportional to the Boltzmann sampling probability, $e^{-E(x)}$ (right plot). Thus, sampling the ESH dynamics over time is equivalent to sampling from the target distribution.}
    \label{fig:rollercoaster}
\end{figure}

We define a separable Hamiltonian, $H(\vx, \vv)$, over position $\vx \in \mathbb R^d$ and velocity $\vv \in \mathbb R^d$ (we use momentum and velocity interchangeably since we set mass to 1). 
$$H(\vx, \vv) = E(\vx) + K(\vv)$$
The potential energy function, $E(\vx)$, is the energy for the target distribution with unknown normalization, $Z$, that we would like to sample, $p(\vx) = e^{-E(\vx)} / Z$, and is defined by our problem. 
We consider an unusual form for the kinetic energy, $K(\vv)$, where $v^2 \equiv \vv \cdot \vv$. 
\begin{equation}
\label{eq:K_ESH}
K_{ESH}(\vv) = \nicefrac{d}{2} \log (v^2/d)    
\end{equation}
For contrast, HMC uses Newtonian kinetic energy, $K (\vv) = \half ~v^2$. 
Hamiltonian dynamics are defined using dot notation for time derivatives and using $\vg(\vx) \equiv \partial_{\vx} E(\vx)$. We also suppress the time dependence, $\vx(t), \vv(t)$, unless needed. 
\begin{align}
&\mbox{General Hamiltonian}&         &\mbox{ESH Dynamics}  &               &\mbox{HMC/Newtonian Dynamics} \nonumber\\
\dot \vx &= \partial_{\vv} H(\vx,\vv)   &      \dot \vx &=  \vv / {\color{red}(v^2/d)}     &      \dot \vx &=  \vv   \label{eq:dynamics} \\
\dot \vv &= -\partial_{\vx} H(\vx,\vv)  &      \dot \vv &= - \vg(\vx)         &      \dot \vv &= - \vg(\vx) \nonumber
\end{align}
Hamiltonian dynamics have a number of useful properties. Most importantly, the Hamiltonian is conserved under the dynamics. Secondly, via Liouville's theorem we see that dynamics are volume preserving (or more generally, \emph{symplectic}). Finally, the dynamics are reversible. While the difference in form between ESH dynamics and Newtonian dynamics appears slight, the effect is significant as we show with an example. 

Imagine riding a roller-coaster running infinitely around a circular track shown in Fig.~\ref{fig:rollercoaster}. $E(x)$ is the potential energy from gravity at a point in the loop specified by a scalar angle $x \in [0, 2 \pi]$. At the bottom, potential energy is small while kinetic energy is large. Under Newtonian physics the speed is slowest at the top so most of a rider's time is spent in dreadful anticipation of the fall, before quickly whipping through the bottom.
In both cases, the ``momentum'' is largest at the bottom, but for the ESH dynamics we see a different outcome. Changes in the roller-coaster position, $\dot x$, happen in slow motion at the bottom when momentum is large and then go into fast-forward at the top of the track where the momentum is small. A spectator will see the rider spending most of their time at the bottom, with low potential energy, and the time spent in each region is exactly proportional to the Boltzmann distribution, as we show mathematically in the next section. 

An intriguing connection between Newtonian and ESH dynamics is discussed in Appendix~\ref{app:q}, which mirrors a thermodynamic effect emerging from Tsallis statistics that leads to anomalous diffusion~\cite{zanette1995thermodynamics}.
Note that $v^2=0$ is a singularity in the dynamics in Eq.~\ref{eq:dynamics}. For 1-D dynamics, this could be an issue, as the velocity can never change signs. However, in higher dimensions, dynamics always avoid the singularity (Appendix~\ref{sec:singularity}). 

\subsection{Ergodic ESH Dynamics Sample the Target Distribution}
Assume that we have an unnormalized target (or Gibbs) distribution specified by energy $E(\vx)$ as $p(\vx) = e^{-E(\vx)}/Z$ with an unknown (but finite) normalization constant, $Z$. 
The reason for defining the non-standard Hamiltonian dynamics in Eq.~\ref{eq:dynamics} is that the uniform distribution over all states with fixed energy, $c$, gives us samples from the Gibbs distribution after marginalizing out the velocity.  
\begin{equation}
    \label{eq:canonical}
    p(\vx, \vv) = \delta\big(E(\vx) + K_{ESH}(\vv) - c\big) / Z'  \qquad \rightarrow \qquad p(\vx) = e^{-E(\vx)}/Z 
\end{equation}

The proof is as follows. 
\begin{align*}
    \highlight{p(\vx)} &= \int d\vv ~p(\vx, \vv) = 1/Z' \int {\color{red} d\vv} ~{\color{blue} \delta\left(E(\vx) + d/2 \log v^2/d - c\right)} \\
    &= 1/Z' \int {\color{red} J(\bm \phi)~d\bm \phi~ \rho^{d-1}~d\rho} ~ {\color{blue} \delta\left(E(\vx) + d \log \rho - c -d/2 \log d \right)} \\
    &= 1/Z' \int {\color{red} J(\bm \phi)~d\bm \phi~ \rho^{d-1}~d\rho} ~{\color{blue} \rho/d ~ \delta(\rho - e^{-E(\vx)/d + c/d + \half \log d})} = \highlight{e^{-E(\vx)} / Z} ~~~~\qed
\end{align*}
In the second line, we switch to hyper-spherical coordinates, with $\rho = |\vv|$ and angles combined in $\bm \phi$. In the third line, we use an identity for the Dirac delta that for any smooth function, $h(z)$, with one simple root, $z^*$, we have $\delta(h(z))=\delta(z-z^*)/|h'(z^*)|$~\cite{cuendet2006jarzynski}. Finally we integrate over $\rho$ using the Dirac delta, with the $\phi$ integral contributing constants absorbed into the normalizing constant.

This proof resembles results from molecular dynamics \cite{nose1984molecular}, where an auxiliary scalar ``thermostat'' variable is integrated over to recover a ``canonical ensemble'' in both the coordinates and momentum. In our case, the momentum variables are already auxiliary so we directly use the velocity variables in a role usually filled by the thermostat. The form of the energy is inspired by the log-oscillator \cite{campisi2012logarithmic}. Although log-oscillator thermostats never gained traction in molecular dynamics because they violate several physical properties~\cite{patra2018zeroth}, this is not relevant for machine learning applications. 

Although Eq.~\ref{eq:canonical} gives us the target distribution as the marginal of a uniform distribution over constant energy states, this is not helpful at first glance because it is not obvious how to sample the uniform distribution either. At this point, we invoke the ergodic hypothesis: the dynamics will equally occupy all states of fixed energy over time.  More formally, the (\emph{phase space}) average over accessible states of equal energy is equal to the \emph{time} average under Hamiltonian dynamics for sufficiently long times. 
\begin{align}
    \label{eq:ergodic}
        \mbox{\emph{Ergodic hypothesis}:}\quad \frac{1}{Z'}\int d\vx d\vv~ \delta(H(\vx,\vv)-c)~ h(\vx, \vv)  = \lim_{T\rightarrow \infty} \frac{1}{T} \int_0^T dt~ h(\vx(t), \vv(t))
\end{align}
The importance and frequent appearance of ergodicity in physical and mathematical systems has spurred an entire field of study~\cite{walters2000introduction} with
celebrated results by Birkhoff and von Neumann~\cite{birkhoff1931proof,neumann1932proof,moore2015ergodic} explaining why ergodicity typically holds. Ergodicity for a specific system depends on the details of each system through $E(\vx)$, and the appearance of hidden invariants of the system breaks ergodicity~\cite{martyna1992nose,tupper2005ergodicity,cuendet2006jarzynski}. While we do not expect hidden invariants to emerge in highly nonlinear energy functions specified by neural networks, systems can be empirically probed through fixed point analysis~\cite{martyna1996explicit} or Lyapunov exponents~\cite{patra2015lyapunov}.
Standard MCMC also assumes ergodicity, but the use of stochastic steps typically suffices to ensure ergodicity~\cite{neal2011mcmc}.

The key distinction to MCMC is that our scheme is fully deterministic, and therefore we use ergodic dynamics, rather than randomness in update proposals, to ensure that the sampler explores the entire space. 
As an alternative to relying on ergodicity, we also give a normalizing flow interpretation.

\subsection{Numerical Integration of the Energy Sampling Hamiltonian (ESH) ODE}\label{sec:numerical}
The leapfrog or Stormer-Verlet integrator~\cite{verlet1967computer} is the method of choice for numerically integrating standard Hamiltonian dynamics because the discrete dynamics are explicitly reversible and volume preserving. Additionally, the error of these integrators are $O(\epsilon^3)$ for numerical step size, $\epsilon$. Volume preservation is not guaranteed by off-the-shelf integrators like Runge-Kutta~\cite{yoshida1990construction}. 
The analogue of the leapfrog integrator for our proposed dynamics in Eq.~\ref{eq:dynamics} follows. 
\begin{align}\label{eq:leap}
\vv(t+\epshalf) &=\vv(t) - \epshalf ~ \vg(\vx(t))  & \mbox{Newtonian/HMC} && \mbox{ESH} \nonumber\\
\vx(t+\eps) &= \vx(t) + \eps ~{\color{red} s} ~ \vv(t+\epshalf)  &{\color{red} s=1} && {\color{red}s= d / v^2(t+\epshalf)} \nonumber \\
\vv(t+\eps) &=\vv(t+\epshalf) - \epshalf ~ \vg(\vx(t+\epsilon))
\end{align}
Unfortunately, for ESH the effective step size for $\vx$ can vary dramatically depending on the magnitude of the velocity. This leads to very slow integration with a fixed step size, as we will illustrate in the results. We consider two solutions to this problem: adaptive step-size Runge-Kutta integrators and a leapfrog integrator in transformed coordinates. 
Although Runge-Kutta integrators do not give the same guarantees in terms of approximate conservation of the Hamiltonian as symplectic integrators, the Hamiltonian was stable in experiments (App.~\ref{app:integrate}). Moreover, looking at Eq.~\ref{eq:canonical}, we see that the exact value of the Hamiltonian is irrelevant, so results may be less sensitive to fluctuations. 

\textbf{Transformed ESH ODE}~~~~  The adaptive time-step in the Runge-Kutta ODE solver is strongly correlated to the magnitude of the velocity (App.~\ref{app:integrate}). This suggests that we might be able to make the integration more efficient if we chose a time-rescaling such that the optimal step size was closer to a constant. We chose a new time variable, $\bar t$, so that $dt = d \bar t ~|\vv| /d$, leading to the transformed dynamics $\dot \vx = \vv / |\vv|, \dot \vv = -|\vv|/d ~\vg(\vx)$ (for notational convenience, we omit the bar over $t$ and continue using dot notation to indicate derivative with respect to the scaled time). 
Next, we re-define the variables as follows: $\vu \equiv \vv / |\vv|, r \equiv \log |\vv|$. The transformed ODE follows. 
\begin{align}\label{eq:scaledode}
& \mbox{ESH Dynamics} &      d \bar t &\equiv dt ~d / |\vv|  &   \dot \vx &= \vu \\
\dot \vx &=  \vv / {(v^2/d)} &      \mbox{with}\qquad    \vu &\equiv \vv / |\vv|         & \Longrightarrow \qquad  \qquad  \dot \vu &= -(\mathbb I-\vu \vu^T) \vg(\vx) / d \nonumber \\
\dot \vv &= - \vg(\vx)   &              r & \equiv \log |\vv|           &     \dot r &= - \vu \cdot \vg(\vx) / d \nonumber
\end{align}
Interestingly, the previously troublesome magnitude of the velocity, captured by $r$, plays no role in the dynamics of $\vx$. However, we must still solve for $r$ because at the end of the integration, we need to re-scale the time coordinates back so that the time average in Eq.~\ref{eq:ergodic} can be applied. 
Again, we can solve this ODE with general-purpose methods like Runge-Kutta or with a leapfrog integrator. 

\textbf{Leapfrog integrator for the time-scaled ESH ODE dynamics}~~~~  
Correctly deriving a leapfrog integrator for the time-scaled ODE is nontrivial, but turns out to be well worth the effort. 
The updates for $\vx, \vu$ are below with the full derivation including $r$ update in App.~\ref{app:leapfrog}.
\begin{align}
 \vu(t+\eps/2) &= \vf(\eps/2, \vg(\vx(t)), \vu(t))  & \text{Half step in }\vu      \nonumber  \\
 \vx(t+\eps) &=  \vx(t) + \eps ~\vu(t + \eps / 2)   & \text{Full step in }\vx  \label{eq:scaled_leap} \\
  \vu(t+\eps) &= \vf(\eps/2,\vg(\vx(t+\eps)), \vu(t+\eps/2) ) & \text{Half step in }\vu \nonumber
\end{align}
\begin{gather*}
\mbox{with}~~~\vf(\eps, \vg, \vu) \equiv  \frac{\vu + \ve ~(\sinh{(\eps~\gd)} + \vu \cdot \ve \cosh{(\eps~\gd)} - \vu \cdot \ve)}{\cosh{(\eps~\gd)} + \vu \cdot \ve \sinh{(\eps~\gd)}}  \mbox{   and    }  \ve \equiv -\vg / |\vg|   
\end{gather*}
The update for $\vu$ is towards the direction of gradient descent, $\ve$. If the norm of the gradient is large $\vu \rightarrow \ve$ and the dynamics are like gradient descent. The update form keeps $\vu$ a unit vector. 

\textbf{Ergodic sampling with the transformed ESH ODE}~~~~ The solution we get from iterating Eq.~\ref{eq:scaled_leap} gives us $\vx(\bar t), \vu(\bar t), r(\bar t)$ at discrete steps for the scaled time variable, $\bar t = 0, \eps, 2 \eps,\ldots, \bar T$, where we re-introduce the bar notation to distinguish scaled and un-scaled solutions. We can recover the original time-scale by numerically integrating the scaling relation, $dt = d\bar t ~|\vv(\bar t)|/d$ to get $t(\bar t) = \int_0^{\bar t} d\bar t' |\vv(\bar t') |/d = \int_0^{\bar t} d\bar t' e^{r(\bar t')}/d$. Using this expression, we transform our trajectory in the scaled time coordinates to points in the un-scaled coordinates, $(t, \vx(t), \vv(t))$, except that these values are sampled irregularly in $t$. 
For an arbitrary test function, $h(\vx)$, sampling works by relating target expectations (left) to trajectory expectations (right). 
$$
\mathbb E_{\vx \sim e^{-E(\vx)}/Z}[h(\vx)] \overset{\text{Eq.~\ref{eq:canonical}}}{=}  \mathbb E_{\vx \sim p(\vx, \vv)}[h(\vx)] \overset{\text{Eq.~\ref{eq:ergodic}}}{\approx} \mathbb E_{t \sim \mathcal U[0,T]} [h(\vx(t))]   = \mathbb E_{\bar t \sim \mathcal U[0,\bar T]} [\vv(\bar t) /D~  h(\vx(\bar t))] 
$$
The approximation is from using a finite $T$ rather than the large $T$ limit. The penultimate form justifies one procedure for ergodic sampling \textemdash~we uniformly randomly choose $t \sim \mathcal U[0,T]$, then take $\vx(t)$ as our sample. Because the time re-scaling might not return a sample that falls exactly at $t$, we have to interpolate between grid points to find $\vx(t)$. Alternatively, the last expression is a weighted sample (by $|\vv|$) in the time-scaled coordinates. We can avoid storing the whole trajectory and then sampling at the end with reservoir sampling~\cite{vitter1985random}. See App.~\ref{sec:algorithm} for details.

\subsection{Alternative Interpretation: Jarzynski Sampling with ESH as a Normalizing Flow}
We can avoid the assumption of ergodicity by interpreting the ESH dynamics as a normalizing flow. 
The idea of interpreting Hamiltonian dynamics as a normalizing flow that can be weighted to sample a target distribution goes back to Radford Neal's unpublished~\cite{neal2005hamiltonian} but influential~\cite{hamiltonianvae,salimans2015markov,rezende2015variational} work on Hamiltonian Importance Sampling. Jarzynski's equality~\cite{jarzynski1997,jarzynski2011} can be related to importance sampling~\cite{habeck} and has been applied to Hamiltonian dynamics~\cite{jarzynski2011,cuendet2006jarzynski} which motivates our approach. Neal's Hamiltonian importance sampling required an annealing schedule to move the initial high energy Newtonian dynamics closer to the low energy target distribution. This is not required for ESH dynamics as our Hamiltonian directly samples the target. 

We initialize our normalizing flow as $\vx(0), \vv(0) \sim q_0(\vx, \vv) = e^{-E_0(\vx)} / Z_0  ~\delta(|\vv|-1)/A_d$, where $E_0$ is taken to be a simple energy model to sample with a tractable partition function, $Z_0$, like a unit normal. Then we transform the distribution using deterministic, invertible ESH dynamics to $q_t(\vx(t), \vv(t))$. The change of variables formula for the ODE in Eq.~\ref{eq:scaledode} (App.~\ref{app:jar})  is 
$$\log q_t(\vx(t), \vv(t)) = \log q_0(\vx(0), \vv(0)) + \log |\vv(0)| - \log |\vv(t)|.$$ 
Then, using importance sampling, we can relate $q_t$ to the target distribution $p(\vx) = e^{-E(\vx)}/Z$.
\begin{gather}\label{eq:jar}
 \mathbb {\color{red} \mathbb E}_{\color{red} q_t(\vx(t), \vv(t))} [e^{{\color{red} w}(\vx(t),\vv(t))}~ h(\vx(t))] = \frac{Z}{Z_0} {\color{red} \mathbb E}_{\color{red} p(\vx)} [h(\vx)] \\
 {\small {\color{red} w}(\vx(t),\vv(t)) \equiv E_0(\vx(0)) - E(\vx(0)) + \log |\vv(t)| - \log |\vv(0)| } \nonumber
\end{gather} 
For the weights, we should formally interpret $\vx(0), \vv(0)$ as functions of $\vx(t), \vv(t)$ under the inverse dynamics.  This expression holds for any $h$, including $h(\vx)=1$, which gives us the partition function ratio $\mathbb E_{q_t}[e^w] = Z/Z_0$. We use this relation to replace $Z/Z_0$ when calculating expectations, which is known as self-normalized importance sampling. Note that as in ergodic sampling in the time-scaled coordinates, the weight is proportional to $|\vv(t)|$. We give a full derivation and tests using ESH-Jarzynski flows to estimate partition functions and train EBMs in App.~\ref{app:jar}, but primarily focus on ergodic sampling results in the rest of the paper.

\section{Results}\label{sec:results} 
ESH dynamics provide a fundamentally different way to sample distributions, so we would like experiments to build intuition about how the approach compares to other sampling methods, before considering applications like sampling neural network energy functions. 
We introduce some standard synthetic benchmarks along with variations to help contrast methods.
\begin{itemize}
    \item \emph{Mixture of Gaussian} (2D MOG) is a mixture of 8 Gaussians with separated modes. 
    \item \emph{Mixture starting with informative prior} (2D MOG-prior) is the same as 2D MOG but initializes samplers from a mode of the distribution. This simulates the effect of using ``informative priors''\cite{nijkamp2} to initialize MCMC chains, as in persistent contrastive divergence (PCD)~\cite{pcd}. The danger of informative initialization is that samplers can get stuck and miss other modes.  
    \item \emph{Ill Conditioned Gaussian} (50D ICG) from \cite{neal2011mcmc} has different length scales in different dimensions.
    \item \emph{Strongly Correlated Gaussian} (2D SCG) with a Pearson correlation of $0.99$.
    \item \emph{Strongly Correlated Gaussian with bias} (2D SCG-bias) is the same as 2D-SCG except we bias the initialization toward one end of the long narrow distribution (as in Fig.~\ref{fig:valley}). This tests the ability of samplers to navigate long, low energy chasms.  
    \item \emph{Funnel} (20D Funnel) A challenging test case where the length scale in one region is exponentially smaller than other regions. Our implementation is from~\cite{proposal_entropy}. 
\end{itemize}
\textbf{Metrics}~~~~ Our goal is to obtain high quality samples from the target distribution. 
With the rise of multi-core CPUs, GPUs, and cloud computing, most practitioners prefer to simulate many Markov chains in parallel for the minimum amount of time, rather than to simulate one chain for a very long time~\cite{stein,nijkamp1}. Therefore, we imagine running a number of samplers in parallel, and then we measure how well they resemble the target distribution after some time, as a function of \emph{number of serial gradient evaluations} per chain. For neural network energy functions, gradient evaluation dominates the computational cost. To measure how well the samples resemble true samples from the distribution we use the (unbiased) kernel estimator of the squared \textbf{Maximum Mean Discrepancy (MMD)}~\cite{gretton2012kernel}, with Gaussian kernel with bandwidth chosen via the median heuristic. MMD is easy to compute with minimal assumptions, converges to zero if the sampling distributions are indistinguishable, and has high statistical power. 
In line with previous work, we also calculate the popular \textbf{Effective Sample Size (ESS)} using the method in~\cite{nuts}.
We caution readers that ESS is a measure of the variance of our estimate \emph{assuming} we have sampled the Markov chain long enough to get unbiased samples. 
For this reason, many MCMC samplers throw away thousands of initial steps in a ``burn-in'' period. 
While we perform our experiments without burn-in in the transient regime, ESS can be interpreted as a proxy for mixing speed, as it depends on the auto-correlation in a chain. 
Cyclic dynamics would violate ergodicity and lead to high auto-correlation and a low ESS, so this can be understood as evidence for ergodicity. Additional empirical evidence of ergodicity is given in Sec.~\ref{sec:ergodic}. 

\begin{figure}[t]
    \centering
    \begin{tabular}{c c c}
    \includegraphics[width=0.33\textwidth,trim={2cm 1cm 2cm 5cm},clip]{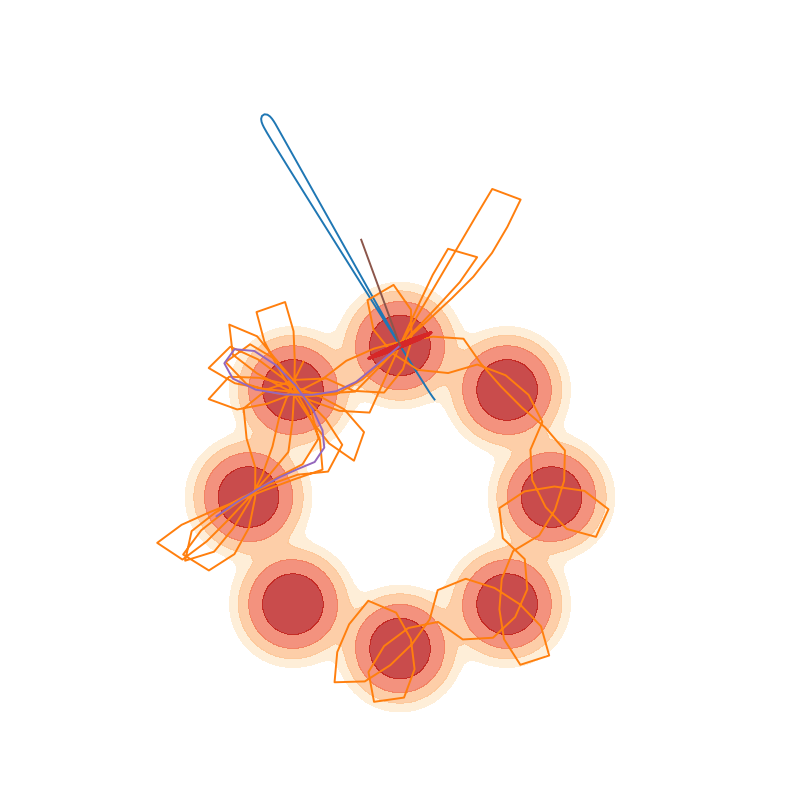} &
    \includegraphics[width=0.4\textwidth,trim={0 0 7cm 0},clip]{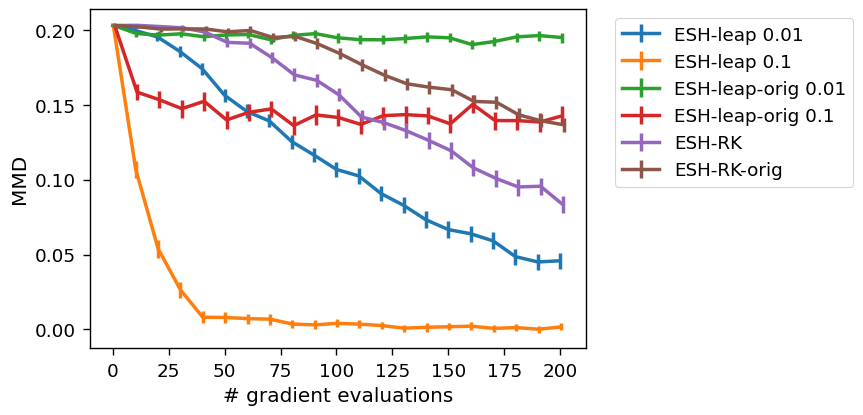} &
    \includegraphics[width=0.25\textwidth,trim={15.5cm 5cm 0 0},clip]{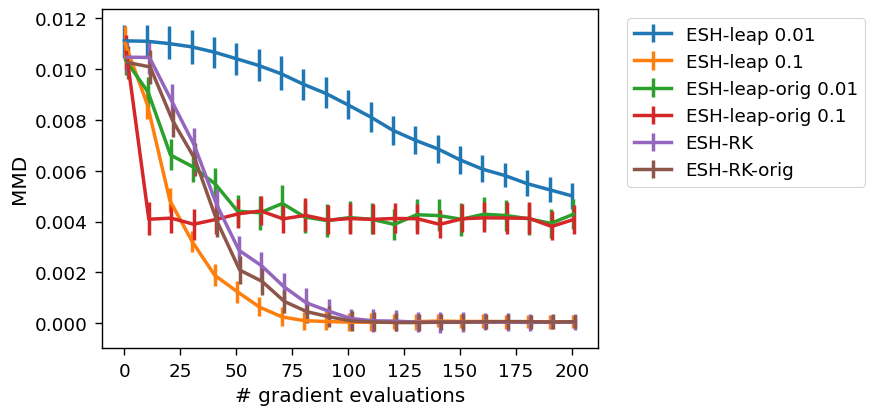}  %
    \end{tabular}
    \caption{(Left) A single chain integrated with 200 gradient evaluations with various ESH ODE solvers on the 2D MOG energy function. (Right) Maximum Mean Discrepancy (MMD) as a function of the number of gradient evaluations. Solutions to the original, unscaled ODE are labeled \texttt{orig}.}
    \label{fig:mmd_esh}
\end{figure}

\subsection{Comparing ESH Integrators} 

For ESH dynamics, we compare a variety of methods to solve the dynamics. First of all, we consider solving the original dynamics (\texttt{orig}), Eq.~\ref{eq:dynamics}, versus the time-scaled dynamics, Eq.~\ref{eq:scaledode}. For each dynamics, we compare using an adaptive Runge-Kutta (\texttt{RK}) solver~\cite{chen2018neural} (fifth order Dormand-Prince~\cite{dormand}) to the leapfrog solvers (\texttt{leap}) in Eq.~\ref{eq:leap} and \ref{eq:scaled_leap} respectively.  In Fig.~\ref{fig:mmd_esh} the scaled dynamics are preferable to the original and the leapfrog integrator is preferable to Runge-Kutta. 
Sec.~\ref{app:integrate} confirms that the leapfrog integrator for the scaled ODE is the best approach across datasets.

For leapfrog integrators, there is only one hyper-parameter, the step size, and for Runge-Kutta integrators we need to choose the tolerance. In experiments we set these to be as large as possible on a logarithmic grid without leading to numerical errors. This leads us to use a value of $\eps=0.1$ in subsequent experiments. For the Runge-Kutta integrators we were forced to use rather small sizes for the relative tolerance ($10^{-5}$) and absolute tolerance ($10^{-6}$) to avoid numerical errors. Since the Hamiltonian should be conserved, it is a good measure of a solver's stability. We plot the Hamiltonian for different solvers in App.~\ref{app:integrate}. Monitoring the error in the Hamiltonian could potentially make an effective scheme for adaptive updating of the step size. The scaled solver makes a step of fixed distance in the input space, but for distributions with varying length scales, this may be undesirable.

\subsection{Comparing ESH with Other Samplers}

\begin{figure}
    \centering
        \begin{tabular}{c c c}
        \tiny \textbf{2D MOG} & \tiny \textbf{2D MOG-Prior} & \tiny \textbf{2D SCG} \\
    \includegraphics[width=0.3\textwidth,trim={0 0 5cm 0},clip]{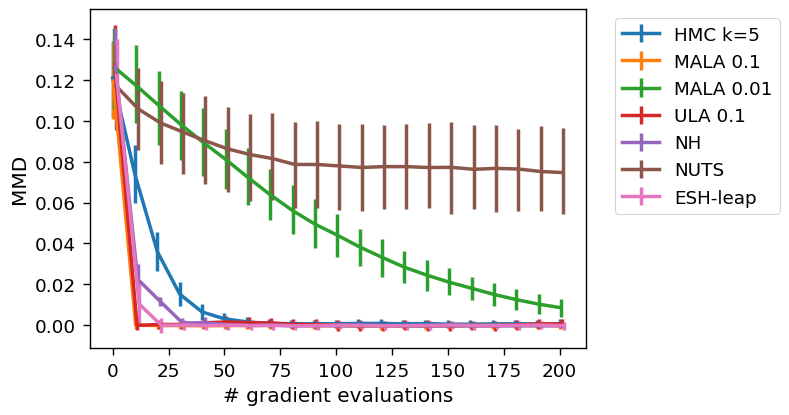} &
    \includegraphics[width=0.3\textwidth,trim={0 0 5cm 0},clip]{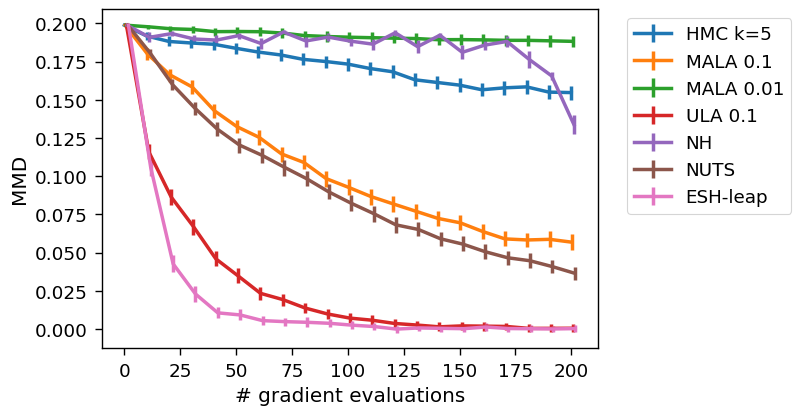} &
    \includegraphics[width=0.3\textwidth,trim={0 0 5cm 0},clip]{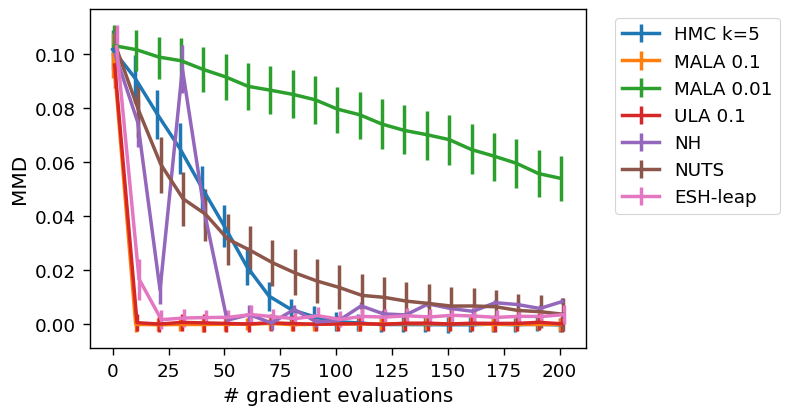} \\
          \tiny \textbf{2D SCG-Bias} & \tiny \textbf{50D ICG} &  \\
    \includegraphics[width=0.3\textwidth,trim={0 0 5cm 0},clip]{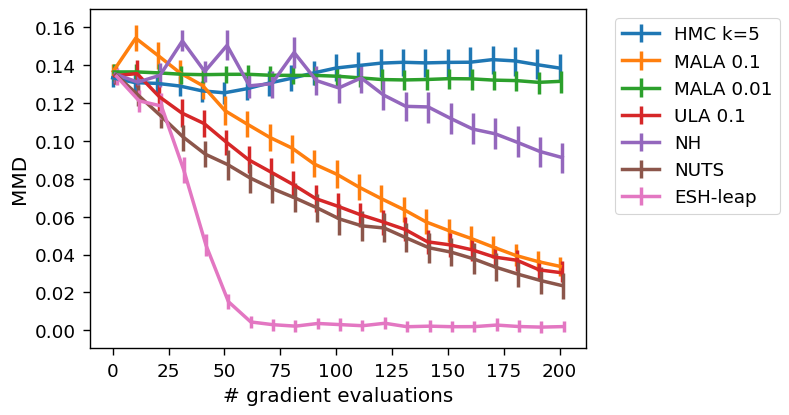} &
    \includegraphics[width=0.3\textwidth,trim={0 0 5cm 0},clip]{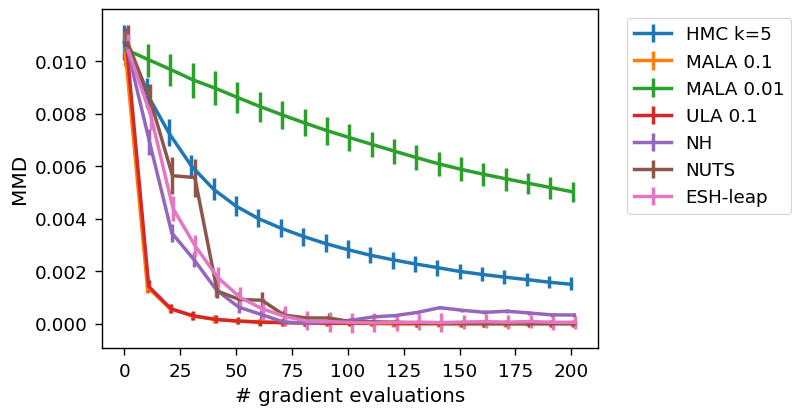} &
    \includegraphics[width=0.18\textwidth,trim={15.5cm 4.5cm 0 0.2cm},clip]{figures/comparison/mmd_50D-ICG.png}
    \end{tabular}
    \caption{Maximum Mean Discrepancy (MMD) as a function of the number of gradient evaluations performed for different samplers.}
    \label{fig:mmd}
\end{figure}

\begin{figure}
    \centering
    \includegraphics[width=0.48\textwidth,trim={0 3.45cm 0 3.2cm},clip]{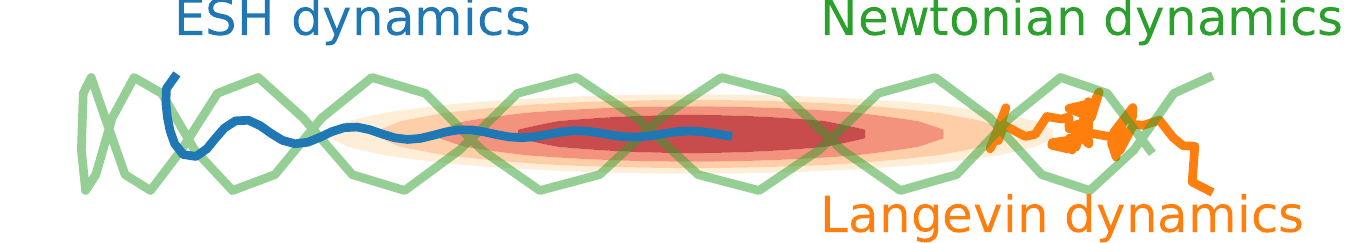}
        \includegraphics[width=0.48\textwidth,trim={0 3.45cm 0 3.2cm},clip]{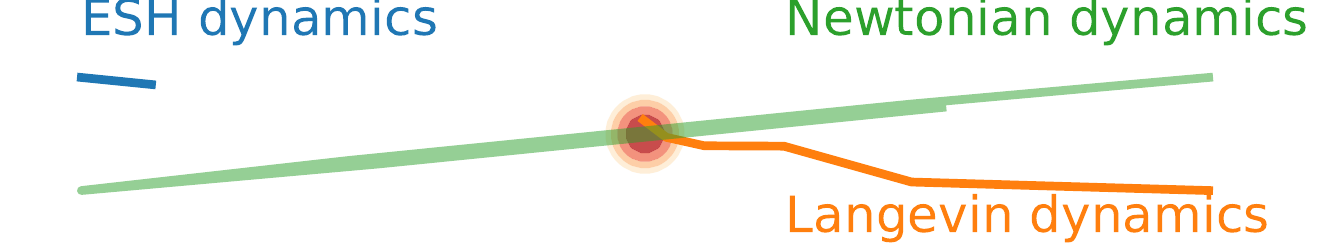}
    \caption{(Left) Navigating long energy valleys with 50 gradient evaluations. (Right) Entering deep energy wells with 5 gradient evaluations.}
    \label{fig:valley}
\end{figure}

We compare with the following sampling methods in our experiments that also use gradient information to sample from energy models. 
\begin{itemize}
    \item \emph{Metropolis-Adjusted Langevin Algorithm} (MALA)~\cite{mala} is a popular sampling method that uses gradient information in the Langevin step, along with a Metropolis-Hastings rejection step to ensure convergence to the target distribution.  
    \item \emph{Unadjusted Langevin Algorithm} (ULA) skips the Metropolis-Hastings rejection step, with the argument that if the step size is made small enough, no rejections will be necessary \cite{sgld}. 
    \item \emph{Hamiltonian Monte Carlo} (HMC)~\cite{neal2011mcmc} uses Hamiltonian dynamics to propose Markov steps that traverse a large distance while still having a high likelihood of being accepted. In the experiments we use $k=5$ steps to see if it improves over MALA/ULA. If more steps are beneficial, we expect this to be discovered by automatic hyper-parameter selection using NUTS. 
    \item \emph{No-U-Turn Sampler} (NUTS)~\cite{nuts}  is a version of HMC with automatic hyper-parameter selection.  
    \item \emph{Nos\'e-Hoover Thermostat} (NH) is a deterministic dynamics in an extended state space that introduces a ``thermostat'' variable in addition to velocity. Based on the original works of Nos\'e~\cite{nose1984molecular} and Hoover \cite{hoover1985canonical}, we used a general numerical integration scheme proposed by Martyna et al. \cite{martyna1996explicit} with the specific form taken from~\cite{kleinerman2008implementations}.  
\end{itemize}
Results are summarized in Fig.~\ref{fig:mmd} and Table~\ref{tab:ess}, with a visualization in Fig.~\ref{fig:esh_movie2}. 
Langevin dynamics give a strong baseline. Nos\'e-Hoover and HMC give mixed results, depending on the energy. While automatic hyper-parameter selection with NUTS is useful in principle, the overhead in gradient computation makes this approach uncompetitive in several cases. ESH is competitive in all cases and a clear favorite in certain situations. Fig.~\ref{fig:valley} illustrates why ESH is particularly effective for the biased initialization examples. Langevin dynamics have a hard time navigating long low energy chasms quickly because of random walk behavior. On the other hand, far from deep energy wells with large gradients, Langevin makes large steps, while ESH is limited to constant length steps in the input space. For this reason, ESH slightly lags Langevin in the ill-conditioned and strongly correlated Gaussians. While we may get the best of both worlds by smartly initializing ESH chains, we investigate only the basic algorithm here. The large scores for effective sample size in Table~\ref{tab:ess} show that ESH dynamics mix quickly. All methods, including ESH, failed on the Funnel example, see App.~\ref{app:funnel}. 

\begin{table}\label{tab:ess}
\tiny
  \caption{Effective Sample Size (with Standard Deviation)}
\begin{tabular}{l|llllll}
\toprule
Sampler &            \textbf{ESH-leap} (Ours) &             HMC k=5 &               MALA 0.1 &                  NH &                NUTS &             ULA 0.1 \\
Dataset      &                     &                         &                     &                     &                     &                     \\
\midrule
20D Funnel   &  \textbf{1.0e-03} (3.0e-04) &  9.6e-04 (1.6e-04)  &  8.8e-04 (1.1e-04) &  9.4e-04 (1.5e-04) &  \textbf{1.0e-03} (1.8e-04) &  8.8e-04 (1.1e-04) \\
2D MOG       &  \textbf{2.1e-02} (1.6e-02) &  2.5e-03 (1.6e-03)  &  4.1e-03 (1.3e-03) &  3.6e-03 (1.3e-03) &  2.2e-03 (2.8e-03) &  8.8e-03 (4.6e-03) \\
2D MOG-prior &  \textbf{2.6e-02} (1.8e-02) &  3.0e-03 (8.4e-04)  &  4.2e-03 (1.5e-03) &  2.7e-03 (3.8e-04) &  4.6e-03 (2.0e-03) &  8.5e-03 (4.5e-03) \\
2D SCG       &  \textbf{2.4e-02} (1.4e-02) &  7.6e-03 (4.8e-03)  &  1.3e-02 (8.0e-03) &  1.0e-02 (5.7e-03) &  1.2e-02 (1.1e-02) &  1.3e-02 (8.1e-03) \\
2D SCG-bias  &  \textbf{8.9e-03} (1.2e-02) &  9.6e-04 (2.1e-03)  &  2.9e-03 (5.1e-03) &  1.8e-03 (3.5e-03) &  2.5e-03 (4.9e-03) &  3.7e-03 (6.0e-03) \\
50D ICG      &  1.6e-04 (2.6e-04) &  2.8e-05 (4.0e-05)  &  7.4e-04 (5.0e-04) &  1.8e-04 (2.2e-04) &  1.2e-04 (1.4e-04) &  \textbf{7.8e-04} (5.1e-04) \\
\bottomrule
\end{tabular}
\end{table}

\subsection{Sampling from Neural Network Energy Models}\label{sec:jem}
\begin{figure}[tbp]
    \centering
    \begin{minipage}{.57\textwidth}
        \includegraphics[width=0.99\textwidth]{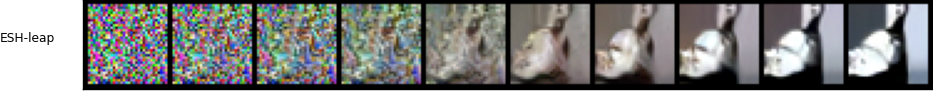}
            \includegraphics[width=0.99\textwidth]{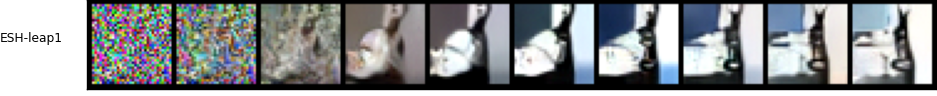}
                                        \includegraphics[width=0.99\textwidth]{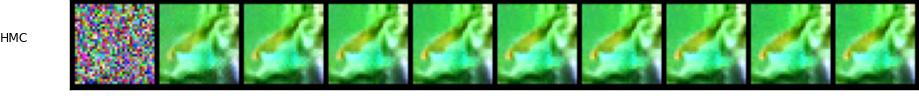}
                    \includegraphics[width=0.99\textwidth]{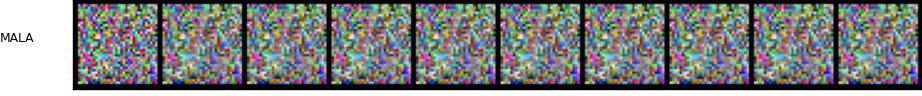}
                        \includegraphics[width=0.99\textwidth]{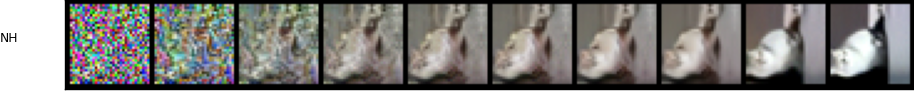}
                                                \includegraphics[width=0.99\textwidth]{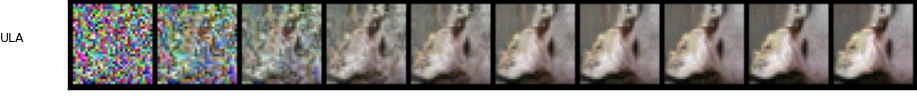}
    \end{minipage}
    \begin{minipage}{0.4\textwidth}
        \includegraphics[width=\textwidth]{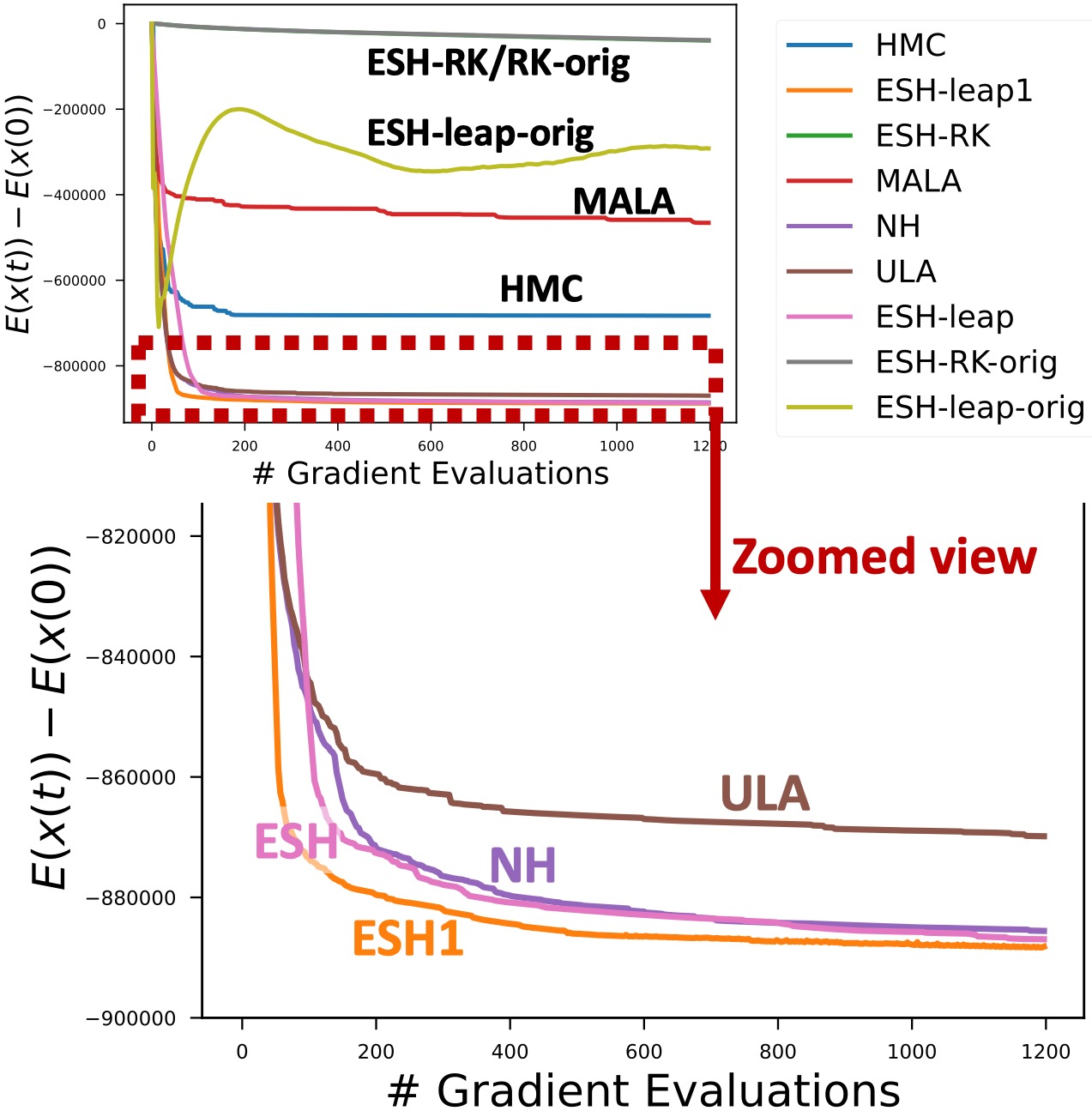} 
    \end{minipage}
    \caption{(Left) Example of sampling chains from random initialization with 200 gradient evaluations per method. (Right) Average energy over time for a batch of 50 samples using different samplers.}
    \label{fig:jem}
\end{figure}

We consider sampling from a pre-trained neural network energy model, JEM~\cite{secret_classifier}. 
Fig.~\ref{fig:jem} shows example chains for samplers starting from random initialization, and Fig.~\ref{fig:prior_jem} shows samplers initialized from a replay buffer constructed during training. We also plot the average energy for a batch of examples. In both cases, ESH finds much lower energy samples than the algorithm used for training, ULA. Nos\'e-Hoover makes a surprisingly strong showing considering its poor performance on several synthetic datasets. We included an ESH leapfrog integrator with a larger step size of $\eps=1$, which performed the best, especially with few gradient evaluations. HMC and MALA do poorly using noise initialization because most proposal steps are rejected. This is because JEM and similar works implicitly scale up the energy in order to take larger Langevin gradient steps that overcome random walk noise~\cite{du,secret_classifier,nijkamp2} as described in \cite{nijkamp1}. Those papers use the common argument that ULA approximates MALA for small step size~\cite{sgld}, but with the large energy scaling employed this argument becomes dubious, as demonstrated by the large gap between ULA and MALA. The large energy scale is introduced to reduce random walk behavior, but ESH dynamics completely avoid random walks.

\subsection{Training Neural Network Energy-Based Models}\label{sec:train}\label{sec:cifar}

Next, we want to compare the quality of results for training neural network energy models using the current state-of-the-art methods which all use Langevin dynamics~\cite{xie2016theory,song2021train,du2020improved,secret_classifier,nijkamp1} to the exact same training using ESH dynamics for sampling. In all cases, an informative prior is used for sampling, persistent contrastive divergence(PCD)~\cite{cd,pcd,du2020improved}. Our synthetic results suggest ESH can be particularly beneficial with informative priors. 
The justification for persistent initialization in MCMC is that the target distribution is invariant under the dynamics, so if the buffer samples have converged to the target distribution then the dynamics will also produce samples from the target distribution. 
We show in App.~\ref{app:stationary} that ESH dynamics also leave the target distribution invariant, justifying the use of PCD. 

For our experiment, we used a setting resulting from the extensive hyper-parameter search in \cite{nijkamp1}. We train a convolutional neural network energy model on CIFAR-10 using a persistent buffer of 10,000 images. 
For each batch, we initialize our samplers with random momentum and image samples from the buffer, then sample using either Langevin or ESH dynamics for 100 steps, and then replace the initial image samples in the buffer. 
Following prior work~\cite{du} we also used spectral normalization~\cite{sngan} for training stability and used ensembling by averaging energy over the last ten epoch checkpoints at test time.
Hyper-parameter details for training and testing are in Sec.~\ref{sec:hyper}. 
For efficient ergodic sampling with the ESH dynamics, we do not store the entire trajectory but instead use reservoir sampling (Alg.~\ref{alg:res}). 

 \begin{figure}[tbp]
     \centering
     \begin{tabular}{c c}
     \textbf{Persistent buffer}: ULA & \textbf{Persistent buffer}: ESH \\
      \includegraphics[width=0.48\textwidth,trim={0 22.8cm 16.8cm 0},clip]{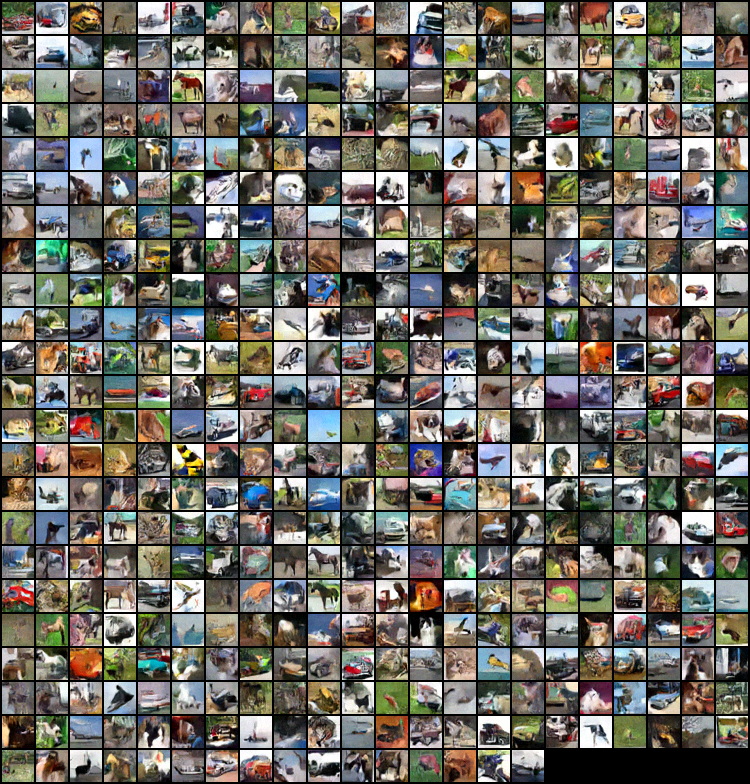} & 
           \includegraphics[width=0.48\textwidth,trim={0 22.8cm 16.8cm 0},clip]{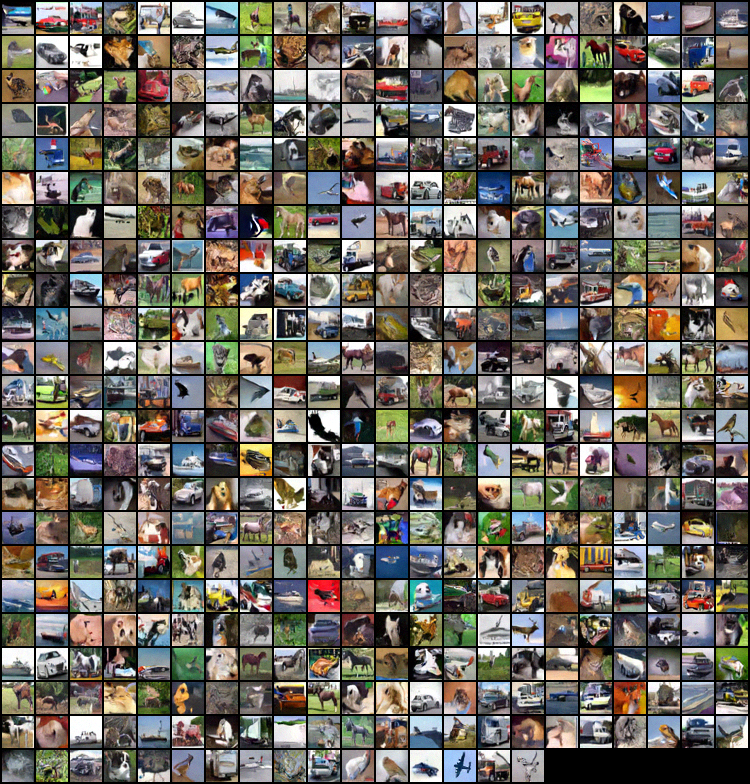} \\
             \textbf{Generate from noise}: ULA & \textbf{Generate from noise}: ESH \\
           \includegraphics[width=0.48\textwidth,trim={0 0 0 0},clip]{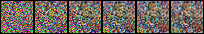} & \includegraphics[width=0.48\textwidth,trim={0 0 0 0},clip]{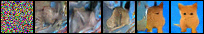} \\
        \textbf{Sampled with random initialization}: ULA & \textbf{Sampled with random initialization}: ESH \\
              \includegraphics[width=0.48\textwidth,trim={0 4.8cm 0 0},clip]{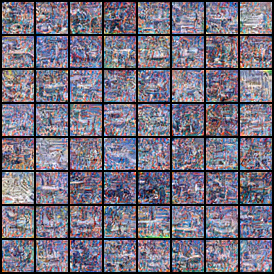} & 
           \includegraphics[width=0.48\textwidth,trim={0 4.8cm 0 0},clip]{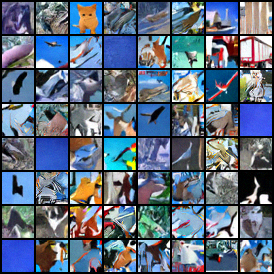}
     \end{tabular}
     \caption{(Top) Samples from the persistent buffer at the end of training. (Middle) Example illustrating sampling from noise. (Bottom) Generated samples initializing from noise and using 15,000 gradient evaluations.}
     \label{fig:cifar}
 \end{figure}
 
Fig.~\ref{fig:cifar} shows that the samples in the persistent buffer produced over the course of training look reasonable for both methods. 
However, this can be misleading and does not necessarily reflect convergence to an energy model that represents the data well~\cite{nijkamp2}.
Generating new samples with chains initialized from noise using 15,000 gradient evaluations per chain reveals a major difference between the learned energy models. The energy model trained with ESH dynamics produces markedly more realistic images than the one trained with Langevin dynamics. 

\begin{figure}[hbtp]
    \centering
    \begin{tabular}{c c c @{\hskip 0.2in} c c c}
    \footnotesize \textbf{True} & \footnotesize \textbf{Train w/ ULA}& \footnotesize \textbf{Train w/ ESH} &     \footnotesize \textbf{True} & \footnotesize \textbf{Train w/ ULA}& \footnotesize \textbf{Train w/ ESH} \vspace{-1mm}\\
     \includegraphics[width=0.14\textwidth,trim={0 6.3cm 11cm 0.65cm},clip]{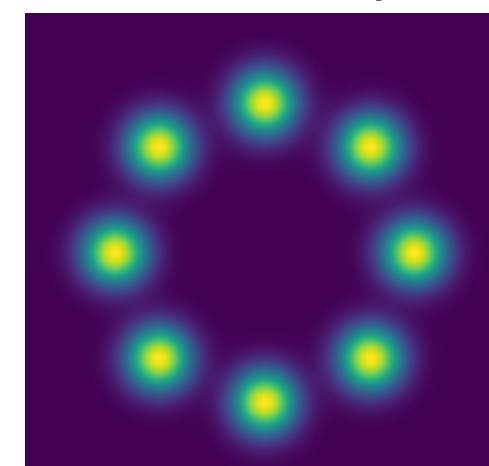} &
     \includegraphics[width=0.14\textwidth,trim={5.5cm 6.3cm 5.5cm 0.65cm},clip]{figures/toy_train/mog_ula_toy_viz_010000.pdf} &
     \includegraphics[width=0.14\textwidth,trim={5.5cm 6.3cm 5.5cm 0.65cm},clip]{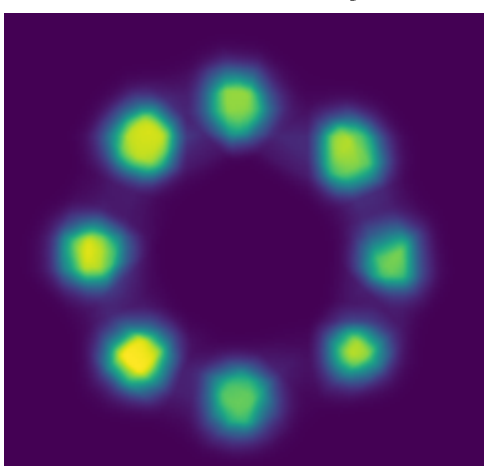} &
     \includegraphics[width=0.14\textwidth,trim={0 6.3cm 11cm 0.65cm},clip]{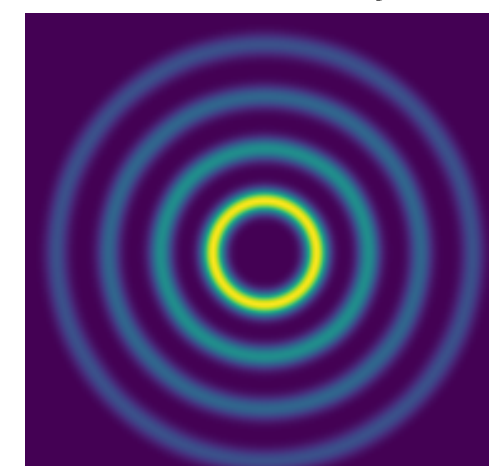} &
     \includegraphics[width=0.14\textwidth,trim={5.5cm 6.3cm 5.5cm 0.65cm},clip]{figures/toy_train/rings_ula_toy_viz_010000.pdf} &
     \includegraphics[width=0.14\textwidth,trim={5.5cm 6.3cm 5.5cm 0.65cm},clip]{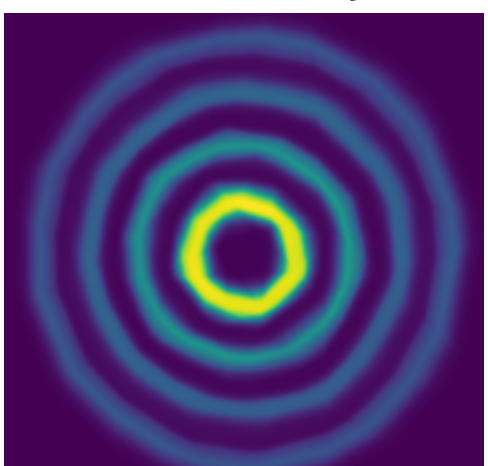}  
    \end{tabular}
    \caption{Trained neural energy models with ULA versus ESH sampling using a total of 500k gradient evaluations over the entire course of training.}\label{fig:toy}
\end{figure}

We also tried the Jarzynski sampler for training energy-based models, with some results on toy data shown in Fig.~\ref{fig:toy}. In this case, the unbiased Jarzynski sampler is very effective at learning to crisply represent boundaries with a small number of total gradient evaluations. However, for higher-dimensional data like CIFAR we found that the higher variance of the sampler becomes problematic. Training details, additional results, and discussion are in App~\ref{sec:jarebm}. 

\section{Related Work} 
Over half a century of research on simulating Molecular Dynamics (MD) has focused on a core issue that also vexes machine learning: how to sample from an energy model. The methods employed in both fields have little overlap because MD is concerned with accurately simulating all physical properties of the target system, including momentum, dynamics of thermalization, and interaction between system and heat bath. Our experiments with Nos\'e-Hoover and other thermostats~\cite{nose1984molecular,hoover1985canonical} often oscillated or converged slowly because the thermostat variable is only weakly coupled to the physical, Newtonian momenta, which are then coupled to the only variables of interest in our case, the original coordinates. A recent review of MD research outlines different methods for constructing thermostats~\cite{md_review} including iso-kinetic thermostats~\cite{morriss1998thermostats} which have some similarities to ESH. 

Ideas from physics like Hamiltonian dynamics and nonequilibrium thermodynamics have inspired many approaches in machine learning~\cite{neal2011mcmc,neal2005hamiltonian,hamiltonianvae,rezende2015variational,habeck,salimans2015markov,ais,jascha_db,jaschaneq,noe_snf}.
Recent work also explores novel ways to combine Hamiltonian dynamics with Jarzynski's equality to derive non-equilibrium samplers~\cite{rotskoff2019dynamical,thin2021invertible}.
Another popular twist on HMC is Riemannian HMC~\cite{rhmc}, where $K(\vv, \vx) = \half \vv^T M(\vx) \vv$ still represents a Newtonian momentum but in curved space. This approach requires second order information like Hessians to define the curvature, and we only considered first order approaches in this paper.  
Another promising line of research recognizes that many of the properties that make Hamiltonian dynamics useful for MCMC come from being part of the class of involutive operators~\cite{involutive} or more general orbits~\cite{orbital}. Another recent line of work also explores sampling via deterministic, continuous dynamics~\cite{neklyudov2021deterministic}.

\section{Conclusion}\label{sec:conclusion}

We presented a new approach for sampling based on deterministic, invertible dynamics and demonstrated its benefits. 
Removing stochasticity leads to faster mixing between modes and could enable applications where backpropagation through the sampler is desired. 
State-of-the-art generative models directly model the score function, the gradient of the energy function,~\cite{song2020score} which can be used to sample with Langevin dynamics but could also potentially benefit from faster mixing with ESH dynamics.
While we introduced an ergodic assumption to motivate the approach, we also provided an intuitive normalizing flow interpretation that does not require ergodicity. 
ESH dynamics provide a simple, fast, and deterministic drop-in replacement for sampling methods like HMC and Langevin dynamics with the potential to impact a wide range of applications such as Bayesian modeling~\cite{izmailov2021bayesian} and latent factor inference~\cite{brekelmans2020all}.

\begin{ack}
We thank Rob Brekelmans for helpful comments on this paper and acknowledge support from the Defense Advanced Research Projects Agency (DARPA) under award FA8750-17-C-0106. We also thank the AWS Cloud Credit for Research Program. 
\end{ack}

\bibliographystyle{unsrt}

\clearpage 

\appendix

\section*{Supplementary Material for ``Hamiltonian Dynamics with Non-Newtonian Momentum for Rapid Sampling''}

\section{Properties of ESH Dynamics}

\subsection{q-Hamiltonian dynamics}\label{app:q}

We can define a continuum of dynamics with Newtonian dynamics at one extreme and energy-sampling dynamics at the other via Tsallis statistics.  
We define a Hamiltonian over position and momentum / velocity variables (we use these terms interchangeably since we set mass to 1), $\vx,\vv \in \mathbb R^d$. 
$$H(\vx, \vv) = E(\vx) + K(\vv)$$
The potential energy function, $E(\vx)$, is the target distribution that we would like to sample and is defined by our problem. 
The kinetic energy, $K$, can be chosen in a variety of ways, but we consider the following class. 
$$K_q(\vv) = d/2 \log_q (v^2/d)$$
We use the $q$-logarithm from Tsallis statistics defined as $log_q(z) = (z^{1-q}-1) / (1-q)$. In the limit $q=1$ we recover the standard logarithm and when $q=0$ the kinetic energy simplifies to the standard, Newtonian, form used in HMC \cite{neal2011mcmc}, $K_0 (\vv) = \half v^2$. 
Hamiltonian dynamics are defined as follows, using the dot notation for time derivatives.
\begin{alignat}{4}
\dot \vx &= \partial H/ \partial\vv &&=  \vv / (v^2/d)^q \\
\dot \vv &= -\partial H/ \partial \vx &&= - \vg(\vx) \equiv - dE/d\vx
\end{alignat}
We did not explore dynamics with intermediate values of $q$, because we only get energy-sampling dynamics for $q=1$. However, we found it interesting that Newtonian and ESH kinetic energy terms could be viewed as opposite ends of a spectrum defined by Tsallis statistics. 

The situation mathematically resembles the thermodynamic continuum that emerges from Tsallis statistics. In that case, standard thermodynamics with Brownian motion corresponds to one extreme for $q$ and in the other extreme so-called ``anomalous diffusion'' occurs~\cite{zanette1995thermodynamics}. Analogously, we can refer to the $q$-logarithmic counterpart of Newtonian momentum as anomalous momentum.

\subsection{The Zero Velocity Singularity}\label{sec:singularity}

The equation $\dot \vx = \vv / v^2 /d$ diverges when $v^2=0$. In 1-d, this is problematic because the sign of $v$ can not switch without going through this singularity. The example in Fig.~\ref{fig:rollercoaster} was 1-d and did give the correct ergodic distribution, but this relied on the fact that the domain was periodic so $v$ never needed to change signs.
For $d>1$, however, this singularity is not an issue. The reason is that conservation of the Hamiltonian by the dynamics prevents $v^2=0$. 
Consider a state, $\vx(0), \vv(0)$, with $H(\vx(0),\vv(0)) =c$. Under the dynamics, we can verify that $\dot H(\vx(t),\vv(t)) = 0$ so that $H(\vx(t),\vv(t))=c$ for all time. 
Therefore, for the ESH Hamiltonian, we have $E(\vx(t)) + d/2 \log v(t)^2 = c$. Then $v(t)^2 = e^{2/d (c-E(\vx(t)))}$, which is positive.

\section{Derivations}

\subsection{Leapfrog Derivation for Scaled Energy Sampling Hamiltonian ODE} \label{app:leapfrog}
We have the following first order ODE. 
\begin{equation}\label{eq:z_ode}
    \dot \vz = \mathcal O \vz
\end{equation} 
This  ODE involves the operator,
$$\mathcal O \equiv \vu \cdot \partial_{\vx} - 1/d~ \vg(\vx)^T (\mathbb I - \vu \vu^T)\partial_{\vu} - 1/d~ \vu \cdot \vg(\vx) \partial_r  
\mbox{   and    }  \vz \equiv (\vx, \vu, r).$$  
Expanding the equation  shows that this equation is equivalent to Eq.~\ref{eq:scaledode}. To be more explicit, $\vz \equiv (x_1,\ldots, x_d, u_1,\ldots u_d, r)$ the operator $\mathcal O$ is a scalar operator that is applied to each element of the vector $\vz$, so we have $2 d +1$ coupled differential equations, $\dot z_i = \mathcal O z_i$. 

The solution to this ODE for small $t$ can then formally be written as $\vz(t) = e^{t \mathcal O} \vz(0)$~\cite{yoshida1990construction}.  This can be seen by simply defining the exponential of an operator in terms of its Taylor series, then confirming this obeys Eq.~\ref{eq:z_ode}. 
The leapfrog comes from the following approximation.
$$e^{\epsilon (\mathcal O_1 +\mathcal O_2)} = e^{\epsilon/2 \mathcal O_2} e^{\epsilon \mathcal O_1} e^{\epsilon/2 \mathcal O_2} + O(\epsilon^3)$$
Note that we can't break up the exponent in the usual way because the operators do not commute. Instead the approximation above is derived using the Baker-Campbell-Hausdorff formula. If we set $\mathcal O_1 = \vu \cdot \partial_{\vx}$ then $\vx(t+\epsilon) = e^{\epsilon \mathcal O_1} \vx(t) = \vx(t) + \epsilon ~\vu(t)$, as in the standard leapfrog formula, Eq.~\ref{eq:leap}. This can be seen by expanding the exponential as a time-series and noting that terms second order and above are zero. 

To derive the updates for $\vu, r$,  we note that the solution to $z(t+\eps) = e^{\epsilon \mathcal O_2} z(t)$ is the same as the solution to the differential equation 
\begin{equation}\label{eq:u}
    \dot \vu = -(\mathbb I-\vu \vu^T) \vg / d \qquad \dot r = - \vu \cdot \vg / d
\end{equation} 
where $\vg \equiv \vg(\vx(t))$ with $\vx$ kept fixed to a constant. Finally, we just need to confirm that the proposed form of the leapfrog updates, below, satisfy this differential equation. \begin{alignat}{2}
    \vu(t+\epsilon) &= \frac{\vu(t) + \ve ~(\sinh{(\eps~\gd)} + \vu(t) \cdot \ve \cosh{(\eps~\gd)} - \vu(t) \cdot \ve)}{\cosh{(\eps~\gd)} + \vu(t) \cdot \ve \sinh{(\eps~\gd)}}      
            &  \qquad \vg &\equiv \vg(\vx(t))  \label{eq:u_leap} \\
    r(t+\epsilon) &= r(t) + \log(\cosh{(\eps~\gd)} + \vu \cdot \ve \sinh{(\eps~\gd)})  \qquad      
            &   \ve &\equiv -\vg / |\vg| \label{eq:r_leap}
\end{alignat}

Consider solving Eq.~\ref{eq:u} with fixed $\vg$ for $\vu(t)$ with initial condition $\vu(0) = \vu_0$. This is a vector version of the Ricatti equation. 
Restating the proposed form of the solution from Eq.~\ref{eq:u_leap} and re-arranging gives
$$
\vu(t) = \frac{\vu_0 - \vu_0 \cdot \ve~ \ve + \ve ~(\sinh{(t~\gd)} + \vu_0 \cdot \ve \cosh{(t~\gd)} )}{\cosh{(t~\gd)} + \vu_0 \cdot \ve \sinh{(t~\gd)}}.
$$
First, note that at $t=0$, we get the correct initial condition. 
Taking the time derivative, we get the following. 
\begin{align*}
    \dot \vu &= d\vu(t)/dt = d/dt \frac{\vu_0 - \vu_0 \cdot \ve~ \ve + \ve ~(\sinh{(t~\gd)} + \vu_0 \cdot \ve \cosh{(t~\gd)} )}{\cosh{(t~\gd)} + \vu_0 \cdot \ve \sinh{(t~\gd)}} \\
             &= -\gd~\vu(t)  \frac{\sinh{(t~\gd)} + \vu_0 \cdot \ve \cosh{(t~\gd)}}{\cosh{(t~\gd)} + \vu_0 \cdot \ve \sinh{(t~\gd)}} + \ve ~\gd~\frac{\cosh{(t~\gd)} + \vu_0 \cdot \ve \sinh{(t~\gd)}}{\cosh{(t~\gd)} + \vu_0 \cdot \ve \sinh{(t~\gd)}} \\
             &=  \vu(t) ~ \vu(t) \cdot \vg /d -\vg/d = -(\mathbb I-\vu(t) \vu(t)^T) \vg / d \qquad\qedsymbol
\end{align*}
We have already used that $- \vu(t) \cdot \vg / d = \frac{\sinh{(t~\gd)} + \vu_0 \cdot \ve \cosh{(t~\gd)}}{\cosh{(t~\gd)} + \vu_0 \cdot \ve \sinh{(t~\gd)}} \gd$. Differentiating $r(t) =\log(\cosh{(t~\gd)} + \vu_0 \cdot \ve \sinh{(t~\gd)}) $ directly recovers the correct differential equation, $\dot r = - \vu \cdot \vg / d$, with initial condition $r(0)=0$.

The full updates including $r$ are as follows.
\begin{align}
 r(t+\eps/2) &= r(t) + a(\eps/2, \vg(\vx(t)))       & \text{Half step in }r \nonumber \\
 \vu(t+\eps/2) &= \vf(\eps/2, \vg(\vx(t)), \vu(t))  & \text{Half step in }\vu      \nonumber  \\
 \vx(t+\eps) &=  \vx(t) + \eps ~\vu(t + \eps / 2)   & \text{Full step in }\vx  \label{eq:full_leap} \\
  \vu(t+\eps) &= \vf(\eps/2,\vg(\vx(t+\eps)), \vu(t+\eps/2) ) & \text{Half step in }\vu \nonumber \\
   r(t+\eps) &= r(t+\eps/2) + a(\eps/2, \vg(\vx(t + \eps)))       & \text{Half step in }r \nonumber 
\end{align}
\begin{align*}
\mbox{with}~~~\vf(\eps, \vg, \vu) &\equiv  \frac{\vu + \ve ~(\sinh{(\eps~\gd)} + \vu \cdot \ve \cosh{(\eps~\gd)} - \vu \cdot \ve)}{\cosh{(\eps~\gd)} + \vu \cdot \ve \sinh{(\eps~\gd)}}     \\
a(\eps, \vg) &\equiv \log(\cosh{(\eps~\gd)} + \vu \cdot \ve \sinh{(\eps~\gd)}) \\
\mbox{and using }~~~  \ve &\equiv -\vg / |\vg|
\end{align*}
As in Eq.~\ref{eq:leap}, the leapfrog consists of a half-step in one set of variables ($\vu, r$), a full step in the other ($\vx$), followed by another half step in the first set. This may appear to require two gradient evaluations per step. However, note that chaining leapfrogs together allows us to re-use the gradient from the previous step, requiring only one gradient evaluation per step.

\subsection{Stationary Distribution of ESH Dynamics}\label{app:stationary}

In the experiments, we used a persistent buffer for initialization. For MCMC, the justification is clear. If the buffer contains samples that are drawn from the target distribution then the transitions, which obey detailed balance, will produce samples that are also drawn from the target distribution. 
While it seems intuitive that starting with samples from the target distribution should be helpful, in this appendix we examine this claim more closely. 
We give an argument based on analysis of stationary distributions for why informative initialization with persistent buffers should also be useful for the scaled ESH dynamics we use in our experiments.

We would like to see how the distribution, $q(\vx, \vv, t)$, changes when we apply the scaled ESH dynamics, assuming that $q(\vx, \vv, 0) = \pi(\vx, \vv) \equiv e^{-E(\vx)} e^{-K(|v|)}/Z$. Here we assume that we have initialized the dynamics with samples from the target distribution and with any spherically symmetric velocities (as the magnitude of the velocity doesn't actually matter for the dynamics of $\vx$). 
The scaled ESH dynamics are $\dot \vx = \vv / |v|, \dot \vv = - |v| \vg(\vx)/d$. We would like to look at how the distribution of samples changes after applying these dynamics. 
$$
\left. \partial_t q(\vx, t) \right|_{t=0}  = \left. \int \partial_t q(\vx, \vv, t) ~dv \right|_{t=0} = - \int ( \nabla_\vx \cdot (\dot \vx~ \pi(\vx, \vv)) +   \nabla_\vv \cdot (\dot \vv~ \pi(\vx, \vv)) )~dv
$$
In the first expression, we expand $q$ in terms of the joint distribution, and in the second we use the continuity equation. We use $dv$ as a shorthand for the volume element for $\vv$, $\partial_t \equiv \partial / \partial t$, and $\nabla_\vx  \cdot \vf \equiv \sum_i \partial_{x_i} f_i$ is the divergence operator. Replacing the time derivatives using the dynamics we get the following. 
\begin{align*}
\partial_t q(\vx, t) &=  -\int ( \nabla_\vx \cdot (\dot \vx~ \pi(\vx, \vv)) +   \nabla_\vv \cdot (\dot \vv~ \pi(\vx, \vv)) )~dv \\ 
&=\int ( \vv / |v| \cdot \vg(\vx) ~\pi(\vx, \vv)  + \pi(\vx, \vv) ~\nabla_\vv \cdot (|v| \vg(\vx)/d) + |v| \vg(\vx)/d \cdot \nabla_\vv \pi(\vx, \vv) )~dv \\
&=\int \pi(\vx, \vv) ( \vv / |v| \cdot \vg(\vx) +  ~ \vg(\vx)/d \cdot \vv / |v| + |v| \vg(\vx)/d \cdot (-K'(|v|) \vv /|v|) )~dv \\
&=\int e^{-E(\vx)} e^{-K(|v|)}/Z ~ \vg(\vx)/d \cdot {\color{red} \vv / |v|} (d +  1  -K'(|v|)|v| ) )~dv \\
&= 0 \qed
\end{align*} 
The first four lines go through straightforward calculations using vector calculus product and chain rules. In the fourth line, we see that the integral over $\vv$ depends almost completely on the magnitude, $|v|$, except for a single linear (anti-symmetric) term that depends on the direction $\vv / |v|$. We can split the integral over the direction into two parts which have the same magnitude but opposite sign, therefore canceling out to give zero. 

What this calculation shows is that if we start with a sample from a buffer which is already converged to the target distribution, and initialize our velocities in a sperically symmetric way, then the density on $\vx$ remains in the target distribution.

\subsection{Objective for Training Energy-based Models}\label{app:train}
\newcommand{\loss}{{\mathcal L}}
To optimize energy-based models, $ p_\theta(x) = e^{-E_\theta(x)} / Z_\theta$, we require gradients like the following. 
\begin{align}
\frac{d}{d\theta} \log p_\theta(x) = - \frac{d E_\theta(x)}{d\theta} - \frac{d \log Z_\theta}{d\theta}
\end{align}
If the energy function is specified by a neural network, the first term presents no obstacle and can be handled by automatic differentiation libraries. The negative log partition function is often called the free energy and its derivative follows. 
\begin{align}\label{eq:dlogZ}
 -\frac{d \log Z_\theta}{d\theta} &= -\frac{1}{Z_\theta}  \frac{d}{d\theta} \int e^{-E_\theta(x)} dx  
 &= \int \frac{e^{-E_\theta(x)}}{Z_\theta} \frac{dE_\theta(x)}{d\theta} dx = \mathbb E_{ p_\theta(x)}\left[\frac{dE_\theta(x)}{d\theta} \right]
\end{align}

For instance, if we were optimizing cross entropy between a data distribution, $p_{d}(x)$, and the distribution represented by our energy model, we would minimize the following loss, $\loss$, whose gradient $d\loss/d\theta$ can be written using Eq.~\ref{eq:dlogZ}.
\begin{align}\label{eq:ce}
\loss &= -\mathbb E_{ p_d(x)} [\log p_\theta(x)] = \mathbb E_{ p_d(x)} [E_\theta(x)] + \log Z_\theta \nonumber
\\
\frac{d \loss}{d\theta} &= \mathbb E_{ p_d(x)} \left[\frac{dE_\theta(x)}{d\theta} \right]  -\mathbb E_{ p_\theta(x)}\left[\frac{dE_\theta(x)}{d\theta} \right] 
\end{align}
This objective requires us to draw (positive) samples from the data distribution and (negative) samples from the model distribution. Intuitively, the objective tries to reduce the energy of the data samples while increasing the energy of negative samples drawn from the model. When the data distribution and the energy model distribution match, the gradient will be zero.

\section{Jarzynski Sampling Derivation and Results}\label{app:jar}

\subsection{Deriving the Jarzynski Sampling Relation}

\paragraph{ESH Dynamics as a Normalizing Flow}
We initialize our normalizing flow as $\vx(0), \vv(0) \sim q_0(\vx, \vv) = e^{-E_0(\vx)}/ Z_0~ q_0(\vv) $, where $E_0$ is taken to be a simple distribution to sample with a tractable partition function, $Z_0$, like a unit normal. The distribution $q_0(\vv) = \delta(|\vv|-1)/A_d$, where $A_d$ is the area of the $d$-sphere. Because of the delta function, this is a normalizing flow on a $2d-1$ dimensional space. 

We transform the distribution using ESH dynamics to $q_t(\vx(t), \vv(t))$, via the deterministic and invertible transformation, $(\vx(t), \vv(t)) = e^{t \mathcal O} (\vx(0), \vv(0)) $, or equivalently via the solution to the following ODE,  
$$\dot \vx = \vv / |\vv|, \dot \vv = -|\vv|/d ~\vg(\vx).$$ 
The distribution $q_t(\vx(t), \vv(t))$ is related to the original distribution via the continuous change of variables formula (or ``phase space compression factor''\cite{cuendet2006jarzynski}) which can be seen as a consequence of the conservation of probability and the divergence theorem or derived as the limit of the discrete change of variables formula~\cite{chen2018neural}.
\begin{align*}
    d/dt ~\log q_t(\vx(t), \vv(t)) &= {\color{red} -\partial_\vx \cdot \dot \vx} - \partial_\vv \cdot \dot \vv \\
    &= {\color{red} 0} +  \partial_\vv \cdot (|\vv|/d ~\vg(\vx)) \\
    &= 1/d ~\vg(\vx) \cdot \vv / |\vv| \\
    d/dt ~ \log |\vv| &= 1/v^2 ~\vv \cdot \dot \vv = - 1/d~ \vg(\vx) \cdot \vv / |\vv|
\end{align*}
The last two lines give us $d/dt ~\log q_t(\vx(t), \vv(t)) = -d/dt ~ \log |\vv(t)|$, and therefore we get the following formula by integrating both sides. 
\begin{equation}
    \label{eq:cov}
\log q_t(\vx(t), \vv(t)) = \log q_0(\vx(0), \vv(0)) +\log |\vv(0)| - \log |\vv(t)|
\end{equation}
This expression is the discrete change of variables for the transformation from $q_0(\vx(0), \vv(0))$ to $q_t(\vx(t), \vv(t))$, and we can now pick out the log-determinant of the Jacobian as $\log |\det \frac{\partial (\vx(0), \vv(0)}{\partial (\vx(t), \vv(t)}| = \log |\vv(0)| - \log |\vv(t)|$. 

\paragraph{Deriving a Jarzynski Equality for ESH Dynamics} 
The essence of the Jarzynski equality~\cite{jarzynski1997}, and non-equilibrium thermodynamics in general, is that distributions for dynamics that are far from equilibrium can be related to the equilibrium distribution through path weighting. Mathematically, the formulation of annealed importance sampling~\cite{ais} is identical, but we prefer the broader motivation of nonequilibrium thermodynamics since our scheme does not have anything that could be viewed as an annealing schedule, and Jarzynski relations have been previously derived for continuous, deterministic dynamics~\cite{cuendet2006jarzynski,jarzynski2011}. 

We begin with the target, ``equilibrium'', distribution that we want to estimate expectations under, $p(\vx) = e^{-E(\vx)}/Z$. The non-equilibrium path is defined by the normalizing flow dynamics above with distribution $q_t(\vx_t, \vv_t)$, where for readability we will distinguish the variables at time $t$ and at time $0$ with subscripts.
\begin{align*}
\mathbb E_{p(\vx_t)}[h(\vx_t)] &= \frac{1}{Z}\int d\vx_t h(\vx_t) e^{-E(\vx_t)} = \frac{1}{Z}\int d\vx_t {\color{red} d\vv_t}  h(\vx_t) e^{-E(\vx_t)} {\color{red} q_t(\vv_t|\vx_t)} \\
&= \frac{1}{Z}\int d\vx_t d\vv_t  h(\vx_t) e^{\color{red} -E(\vx_0) + d \log |\vv_t| - d \log |\vv_0| } q_t(\vv_t|\vx_t)  \quad \mbox{Conservation of Hamiltonian} \\
&= \frac{\color{red} Z_0}{Z}\int d\vx_t d\vv_t  h(\vx_t) e^{\color{red} -E_0(\vx_0)}/ {\color{red}Z_0} e^{{\color{red}E_0(\vx_0)} -E(\vx_0) + d \log |\vv_t| - d \log |\vv_0| } q_t(\vv_t|\vx_t) \\
&= \frac{Z_0}{Z}\int d\vx_t d\vv_t  h(\vx_t) { \color{red} q_0(\vx_0, \vv_0)} e^{E_0(\vx_0) -E(\vx_0) + d \log |\vv_t| - d \log |\vv_0| } \frac{q_t(\vv_t|\vx_t)}{\color{red} q_0(\vv_0)} \\
&= \frac{Z_0}{Z}\int {\color{red} d\vx_0 d\vv_0}  h(\vx_t)  q_0(\vx_0, \vv_0) e^{E_0(\vx_0) -E(\vx_0) + (d+ {\color{red} 1}) (\log |\vv_t| - \log |\vv_0|) } \frac{q_t(\vv_t|\vx_t)}{ q_0(\vv_0)} \\
\end{align*}
In the last line, we apply a change of variables using the determinant of the Jacobian derived for the normalizing flow. Besides multiplying by factors of 1 and re-arranging, the main step was to use the conservation of the Hamiltonian for the dynamics in the second line. The next tricky step is to find an expression for $\frac{q_t(\vv_t|\vx_t)}{ q_0(\vv_0)}$. First, we change the velocity variables to hyper-spherical coordinates, $\rho, \phi$, where the well known formula for the determinant of the Jacobian is $ |\det \frac{\partial \vv}{\partial (\rho, \phi)}| = \rho^{d-1} J(\phi)$.
\begin{align*}
    \frac{q_t(\vv_t|\vx_t)}{ q_0(\vv_0)} &= \frac{q_t(\rho_t, \phi_t|\vx_t) }{ q_0(\rho_0, \phi_0)} \frac{\rho_0^{d-1} J(\phi_0)}{\rho_t^{d-1} J(\phi_t)} & \mbox{Change to spherical} \\
    &= \frac{q_t(\rho_t |\phi_t,\vx_t)  \cancel{J(\phi_t)/A_d}}{ \delta(\rho_0-1) \cancel{J(\phi_0)/A_d} } \frac{\rho_0^{d-1} \cancel{J(\phi_0)}}{\rho_t^{d-1} \cancel{J(\phi_t)}} & \mbox{Expand $q$} \\
    &= \frac{\delta(\rho_t - \rho_0 e^{(E(\vx_0) - E(\vx_t))/d})  }{ \delta(\rho_0-1)} \frac{\rho_0^{d-1}}{\rho_t^{d-1}} & \mbox{Conservation of Hamiltonian} \\
        &=  \frac{\rho_0^{d}}{\rho_t^{d}} =  \frac{|\vv_0|^{d}}{|\vv_t|^{d}} & \mbox{Dirac property + Cons. Hamiltonian} \\
\end{align*}
Now we can plug this expression back into our previous derivation to get the following. 
\begin{align*}
\mathbb E_{p(\vx_t)}[h(\vx_t)] &=\frac{Z_0}{Z}\int  d\vx_0 d\vv_0  h(\vx_t)  q_0(\vx_0, \vv_0) e^{E_0(\vx_0) -E(\vx_0) +  \log |\vv_t| - \log |\vv_0| } \\
\mathbb E_{p(\vx_t)}[h(\vx_t)] &= \frac{Z_0}{Z} \mathbb E_{q_0(\vx_0, \vv_0)}[e^{w(\vx_0, \vv_0)} h(\vx_t)] \\
\quad w(\vx_0, \vv_0) &\equiv E_0(\vx_0) -E(\vx_0) +  \log |\vv_t| - \log |\vv_0|  \qed
\end{align*}
This recovers the expression in Eq.~\ref{eq:jar}. Note that because of the invertible relationship between $\vx_t, \vv_t$ and $\vx_0, \vv_0$ we can view this as an expectation over either set of variables, where the other set is considered a function of the first. We use this expression for sampling in Alg.~\ref{alg:jar}.

\subsection{Testing Jarzynski ESH Sampler by Estimating the Partition Function}

\begin{figure}[htbp]
    \centering
    \includegraphics[width=0.66\textwidth]{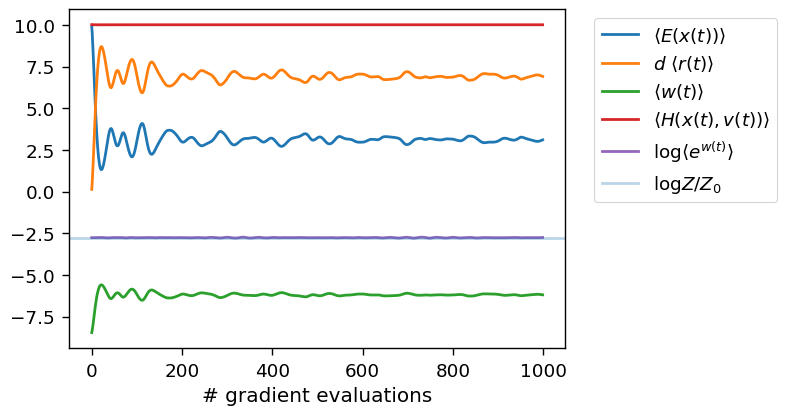}
    \caption{Plotting various quantities for ESH dynamics with a random Gaussian as the energy model.}
    \label{fig:jar}
\end{figure}

To verify this expression, we take a case where the ground truth partition function is known, an energy model for a random Gaussian distribution. Fig.~\ref{fig:jar} plots several quantities, with brackets representing expectations over a batch of trajectories and $E_0(\vx) = 1/2 x^2$. First, we see that the Hamiltonian is approximately conserved. The average energy starts high then quickly converges to a low energy state, while the kinetic energy ($d ~r(t)$) starts low and becomes large. Finally, we see that the weights in Eq.~\ref{eq:jar} correctly give an estimate of the partition function. Note that at $t=0$, the weights correspond to standard importance sampling. 

\subsection{Training Energy-Based Models}\label{sec:jarebm}

Because the Jarzynski sampler is weighted, we cannot estimate the MMD metrics used in Sec.~\ref{sec:results} in the same way. 
Instead, we test the approach by training an energy model to match a distribution with a known ground truth. We consider an energy model, $p(\vx) = e^{-E_\theta(\vx)} / Z$, where $E_\theta(\vx)$ is specified by a neural network. Estimating gradients to train such a model to maximize the likelihood of the data requires sampling from the energy model (App.~\ref{app:train}).  Fig.~\ref{fig:toy} visualizes the energy landscape learned from data sampled from a 2-D distribution. The architecture and hyper-parameters are taken from \cite{nijkamp1} and included in App.~\ref{sec:architecure}. Samplers are initialized with noise and then run for 50 steps per training iteration, using the Jarzynski sampler for ESH.  Better sampling leads to better gradients and therefore the model trained with ESH sampling more crisply matches the true distribution.
\begin{figure}[hbtp]
    \centering
    \begin{minipage}{0.5cm}
         \textbf{Train w/ ESH} \\ ~~ \\ ~~ \\ ~~ 
    \end{minipage} ~~~~
    \includegraphics[width=0.6\textwidth,trim={0 30cm 11cm 0},clip]{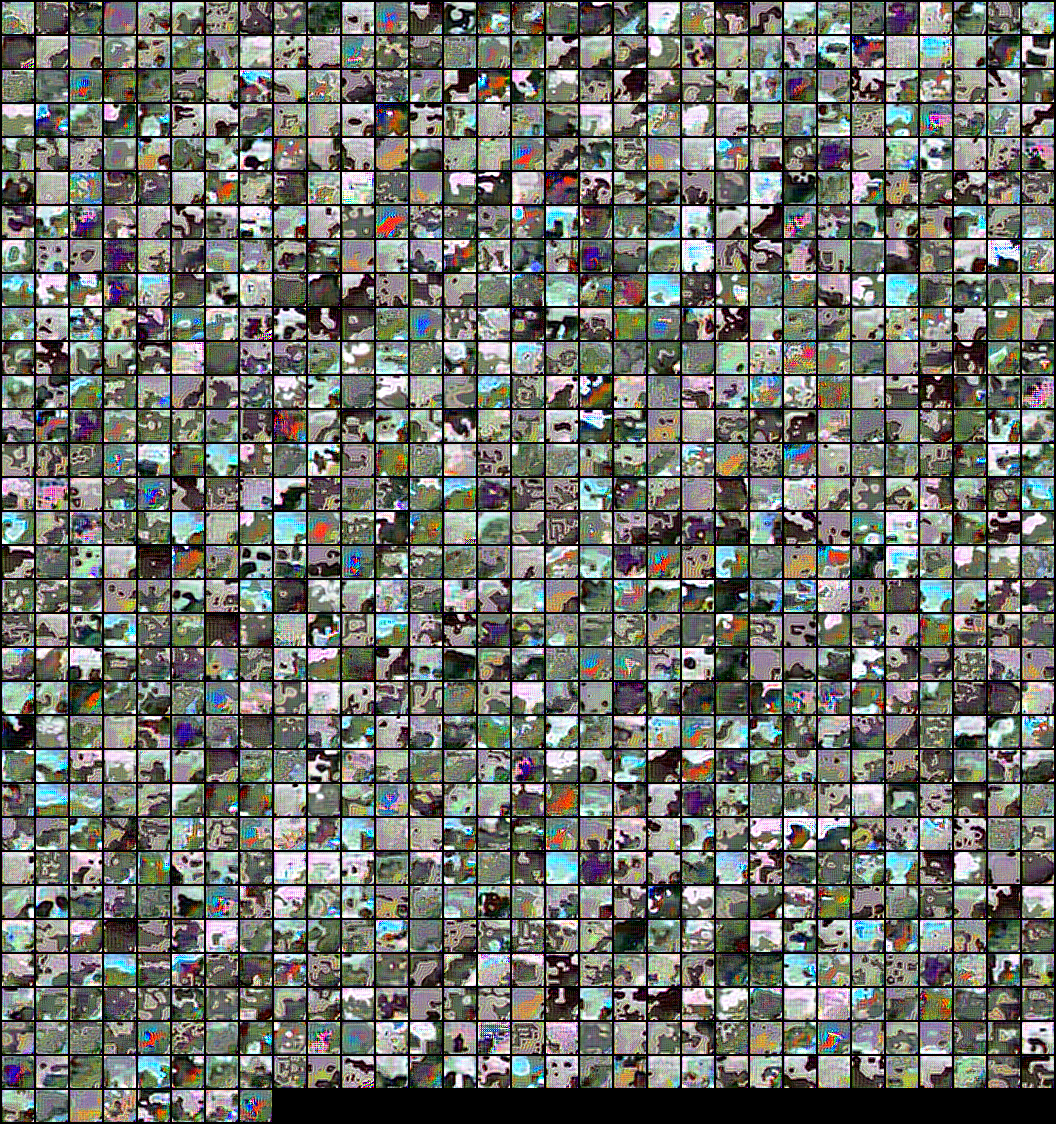} \\
        \begin{minipage}{0.5cm}
         \textbf{Train w/ ULA} \\ ~~ \\ ~~ \\ ~~ 
    \end{minipage}  ~~~~
    \includegraphics[width=0.6\textwidth,trim={0 30cm 11cm 0},clip]{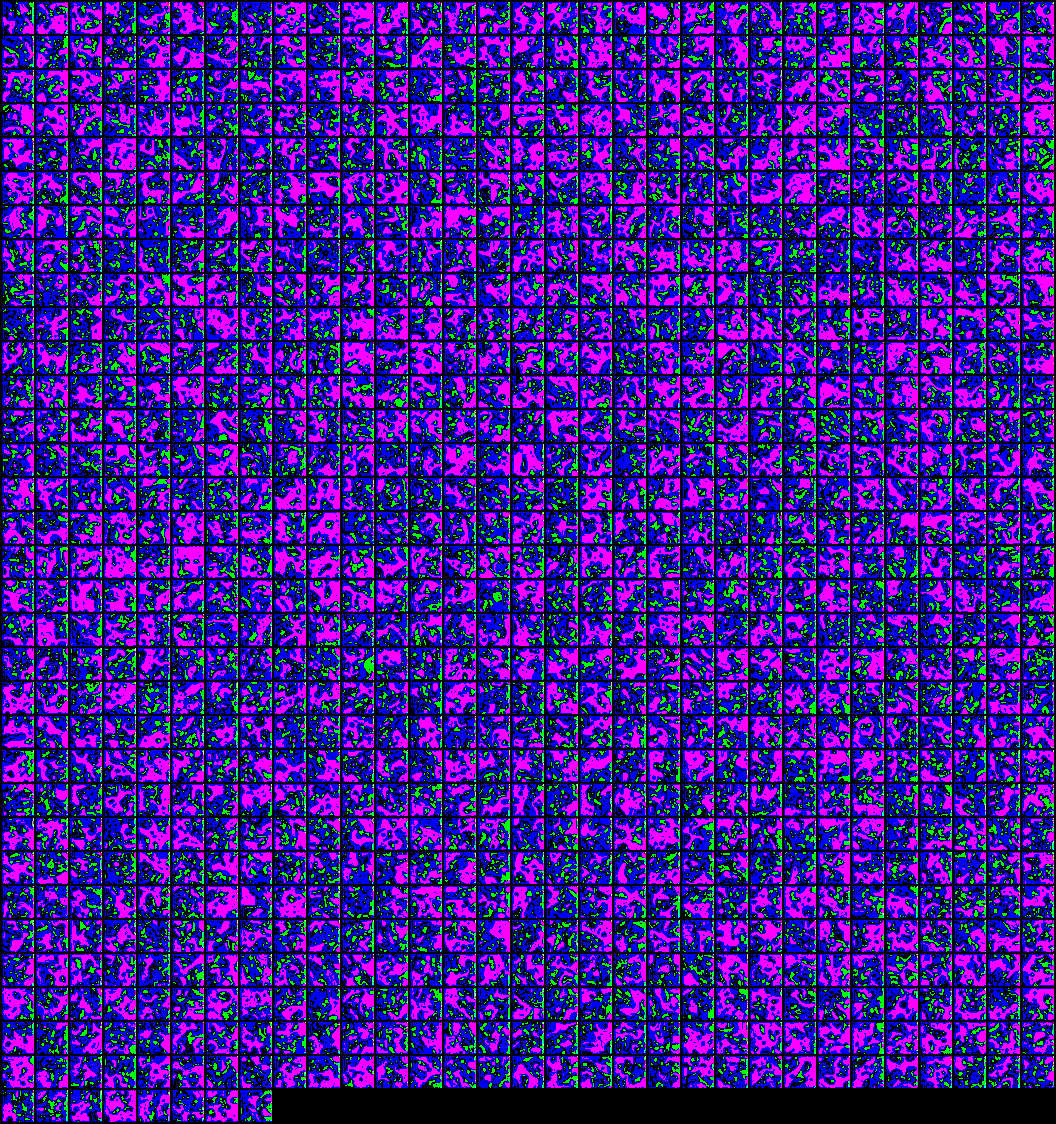}
    \caption{Results training neural energy models using ULA versus ESH sampling with 500k gradient evaluations. (Top) Learned energy for synthetic datasets with noise initialization. (Bottom) Examples from the training buffer using persistent contrastive divergence on CIFAR-10.}\label{fig:nijkamp}
\end{figure}

\paragraph{Unstable Regime for Langevin Dynamics}
Training EBMs with Langevin dynamics for sampling is difficult and requires a great deal of hyper-parameter search to succeed \cite{du,secret_classifier,nijkamp1,nijkamp2,grathwohl2020no}. For the experiment in Fig.~\ref{fig:nijkamp}, we wanted to demonstrate that the unbiased Jarzynski sampler based on ESH dynamics could lead to stable training in a regime where training with Langevin dynamics is unstable. To that end, we considered a large model for CIFAR, with short chains (50 steps), and we omitted heuristics like adding data noise that are typically used to improve stability. We initialized chains using persistent contrastive divergence with a large buffer. All hyper-parameters are listed in Sec.~\ref{sec:hyper}. Predictably, the EBM trained with Langevin dynamics fails egregiously without carefully chosen hyper-parameters. The training with ESH dynamics was stable and led to visually plausible color distributions and diverse signals. However, although training was stable, we noticed that the quality of the solutions stopped improving after the early epochs. We now discuss the reason. 

\begin{figure}[htbp]
    \centering
    \includegraphics[width=0.4\textwidth,trim={11cm 0 0 6cm},clip]{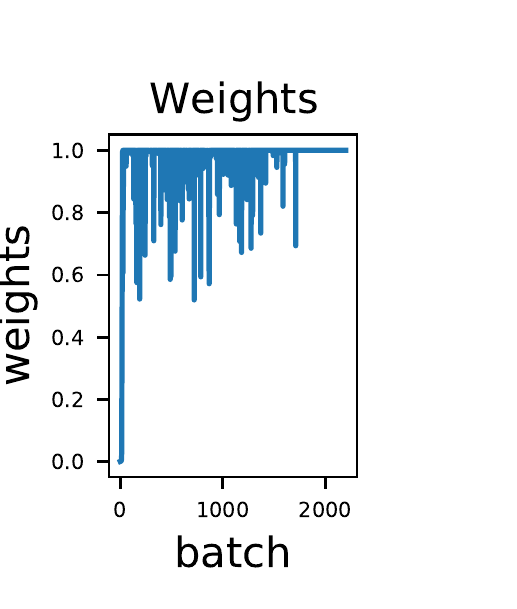}
        \includegraphics[width=0.4\textwidth,trim={11cm 6cm 0 0},clip]{figures/diagnosis_plot.pdf}
    \caption{The maximum weight per batch over time for Jarzynski ESH sampling during CIFAR-10 energy-based model training. We also show the average energy for batches of data versus sampled batches.}
    \label{fig:diagnosis}
\end{figure}

\paragraph{Bias-Variance Trade-off for Jarzynski Sampling} 
Methods like importance sampling use weights to re-weight samples to make one distribution resemble another. A core issue with this approach is that if some weights are much larger than others, then samples that are rare but have very high weight may be missed in a given batch of samples. This increases the variance of the estimator. Although the Jarzynski ESH sampler is an unbiased estimator for expectations under the energy model, it may have very high variance if the weights are very unbalanced. For the 2-d examples in Fig.~\ref{fig:toy} this was not an issue. However, when training high-dimensional energy models for CIFAR, it was an issue. Fig.~\ref{fig:diagnosis} shows the maximum weight per batch for training an EBM for CIFAR. After a few iterations, almost all the weight ends up concentrated on a single image in each batch. This leads to a high variance estimator with noisy gradient estimates. 
In contrast, using persistent contrastive divergence will have higher bias, but because all samples are equally weighted it will also have lower variance.

\section{Algorithms and Implementation}\label{sec:algorithm}

\subsection{Leapfrog Integrator for ESH Dynamics}

The algorithm for the leapfrog integrator of ESH dynamics in the time-scaled coordinates is given in Alg.~\ref{alg:esh}. For implementation, we can improve numerical accuracy by scaling the numerator and denominator of the $\vu$ updates in Eq.~\ref{eq:scaled_leap} by $e^{-2 |\vg| \eps /d}$, so that we are more likely to get underflows than overflows in the exponentials. We also found that we had to treat $\vu \cdot \ve = -1$ as a special case, where the update simplifies to $\vu(t+\eps) = -\ve$. 
A PyTorch implementation is available at \url{https://github.com/gregversteeg/esh_dynamics}.

We simulate in time-scaled coordinates on a regular grid, $\bar t = 0, \eps, 2 \eps,\ldots, \bar T$. 
We can easily convert from scaled to un-scaled time using the relation $dt = d\bar t ~|\vv(\bar t)|/d$ which gives us the following.
$$
t(\bar t) = \int_0^{\bar t} d\bar t' |\vv(\bar t') |/d
$$
We can also convert solutions in terms of $\vv$ or $\vu, r$ using Eq.~\ref{eq:scaledode}.

\begin{figure}[H]
\centering
\noindent\begin{minipage}{0.8\textwidth}
\begin{algorithm}[H]
\begin{algorithmic}[1]
\REQUIRE $E(\vx)$ ~~~~~~~~~~~~~~~~~~~~~~~~~~~~~~~~~~~~~~~~~~~~~~~~~~~~~~~\# Target energy to sample
\REQUIRE $\eps, N$~~~~~~~~~~~~~~~~~~~~~~~~~~~~~~~~~~~~~~~~~~~~~~~~~~~~~~~~~~~~ \# step size, number of steps
\STATE $\bar T=\eps N$ ~~~~~~~~~~~~~~~~~~~~~~~~~~~~~~~~~~~~~~~~~~~~~~~~~~~~~~~~~~~~~~\# total time
\STATE Initialize $\vx(0) \sim \mathcal N(0, I)$ or accept as input 
\STATE $\vu(0) \sim \mathcal N(0,I)$, $r(0)=0$
\STATE $\vu(0) = \vu(0) / |\vu(0)|$
\STATE $\vg(\vx) = \partial_\vx E(\vx)$~~~~~~~~~~~~~~~~~~~~~~~~~~~~~~~~~~~~~~~~~~~~~~~~~\# Gradient of energy
\FOR{$i = 1\dots N$}
\STATE \# Eq.~\ref{eq:full_leap} updates
\STATE $\vu(\bar t+\eps/2) = \vf(\vx(\bar t), \vu(\bar t), \eps/2)$  ~~~~~~~~~~~~~~~~~~~\# Half step $\vu$
\STATE $r(\bar t+\eps/2) = r(\bar t) + a(\eps/2, \vg(\vx(\bar t)))$~~~~~~~~~~~~~\# Half step $r$
\STATE $\vx(\bar t+\eps) = \vx(\bar t) + \eps \vu(\bar t+\eps/2)$~~~~~~~~~~~~~~~~~~~~~~\# Full step in $\vx$
\STATE $\vu(\bar t+\eps) = \vf(\vx(\bar t+\eps), \vu(\bar t+\eps/2), \eps/2)$~~~~~~~\# Half step $\vu$
\STATE $r(\bar t+\eps) =  r(\bar t) + a(\eps/2, \vg(\vx(\bar t+\eps)))$~~~~~~~~~~\# Half step $r$
\ENDFOR
\RETURN $\vx(0),\ldots, \vx(\bar T), \vu(0),\ldots, \vu(\bar T), r(0),\ldots, r(\bar T)$
\end{algorithmic}
\caption{Time scaled leapfrog integrator for ESH dynamics.}
\label{alg:esh}
\end{algorithm}
\end{minipage}
\end{figure}

\subsection{Sampling Algorithms Using ESH Dynamics} 

Recalling from Sec.~\ref{sec:numerical}, we relate expectations under our target distribution to expectations over trajectories using ergodicity. 
$$
\mathbb E_{\vx \sim e^{-E(\vx)/Z}}[h(\vx)] \overset{\text{Eq.~\ref{eq:canonical}}}{=}  \mathbb E_{\vx \sim p(\vx, \vv)}[h(\vx)] \overset{\text{Eq.~\ref{eq:ergodic}}}{\approx} \mathbb E_{t \sim \mathcal U[0,T]} [h(\vx(t))]   
$$
The expression on the right leads to a method for ergodic sampling in Alg.~\ref{alg:ergodic}. 

However, in general we may not want to store the entire trajectory of samples, especially in high-dimensional spaces like images. This prompts us to construct a slightly different estimator. 
\begin{align*}
 \mathbb E_{t \sim \mathcal U[0,T]} [h(\vx(t))]   &= \frac{1}{T} \int_0^T dt~h(\vx(t)) \\
 \mathbb E_{t \sim \mathcal U[0,T]} [h(\vx(t))]   &= \frac{1}{\bar T} \frac{\bar T}{T}  \int_0^{\bar T} d\bar t~|\vv(\bar t)|/d ~h(\vx(\bar t)) \\ 
 &= \mathbb E_{\bar t \sim \mathcal U[0,\bar T]} [\vv(\bar t) /D~  h(\vx(\bar t))] \\
 \text{where}\quad  D &\equiv \frac{1}{\bar T}  \int_0^{\bar T} d\bar t~|\vv(\bar t)|/d = \mathbb E_{\bar t \sim \mathcal U[0,\bar T]} [\vv(\bar t)]
\end{align*}
If we discretize this estimator, we get the following. 
\begin{align*}
     \mathbb E_{t \sim \mathcal U[0,T]} [h(\vx(t))] \approx \sum_{i=0}^N w_i ~h(\vx(\bar t_i))   \quad \text{with} \quad w_i \equiv \frac{ |\vv(\bar t_i)| }{\sum_{j=0}^N |\vv(\bar t_j)|}
\end{align*}
Now we can draw samples according to this normalized weight to sample from the target distribution. Instead of storing the entire sequence of weights and states, $w_i, \vx(\bar t_i)$, we can sample by 
storing a single sample/image in a buffer and updating in an online way by replacing it with the next sample with some probability.  This classic technique is called reservoir sampling~\cite{vitter1985random}. The final state of the buffer will correspond to a state drawn according to the weights, $w$. The procedure is shown in Alg.~\ref{alg:res}.

\hfill
\begin{figure}[H]
\centering
\noindent\begin{minipage}{0.46\textwidth}
\begin{algorithm}[H]
\begin{algorithmic}[1]
\small
\REQUIRE $\vx(\bar t_1),\ldots, \vx(\bar t_N)$ from Alg.~\ref{alg:esh}
\REQUIRE $r(\bar t_1),\ldots, r(\bar t_N)$ from Alg.~\ref{alg:esh}
\FOR{$i=1\dots N$}
\STATE $t_i = \int_0^{\bar t_i} d\bar t' e^{r(\bar t')}/d$
\ENDFOR
\STATE $T = t_N$
\STATE Select $t^*$ randomly from $[0,T]$
\STATE Linearly interpolate to get $\vx^* \equiv \vx(t^*)$
\RETURN $\vx^*$  ~~ \# $\vx^* \sim p(\vx)$ if ergodicity holds
\end{algorithmic}
\caption{Ergodic sampling}
\label{alg:ergodic}
\end{algorithm}
\end{minipage}
\noindent\begin{minipage}{0.52\textwidth}
\begin{algorithm}[H]
\begin{algorithmic}[1]
\small
\REQUIRE $\eps, N, E$~~ \# step size, steps, Energy
\STATE $\vx^* = \emptyset$   ~~~~~\# (Empty) buffer for current sample
\STATE $\text{Cum-weight} = 0$
\FOR{$i=1\dots N$}
\STATE $t, \vx, \vv = \text{Alg}~\ref{alg:esh}(E, \eps, 1)$  \# 1 step of dynamics
\STATE $\text{Cum-weight} = \text{Cum-weight} + |\vv|$
\STATE $\vx^* \leftarrow \vx$ with probability $|\vv| / \text{Cum-weight}$ 
\ENDFOR
\RETURN $\vx^*$ ~~\# sample from $p(\vx)=e^{-E(\vx)}/Z$
\end{algorithmic}
\caption{Reservoir sampling \cite{vitter1985random}}
\label{alg:res}
\end{algorithm}
\end{minipage}
\end{figure}

Jarzynski sampling has a different interpretation based on normalizing flows and importance sampling, as described in Sec.~\ref{app:jar}. The algorithm is summarized in Alg.~\ref{alg:jar}.
\begin{figure}[H]
\centering
\noindent\begin{minipage}{0.95\textwidth}
\begin{algorithm}[H]
\begin{algorithmic}[1]
\small
\REQUIRE $E_0, E$~~ \# Initial energy and target energy
\STATE $(\vx^1(0),\ldots,\vx^n(0)) \sim e^{-E_0(\vx(0))}$ ~~\# Batch size $n$
\STATE $\vx^j(t), \vu^j(t), r^j(t)$ from Alg.~\ref{alg:esh}
\STATE $w^j(t) = E_0(\vx^j(0)) - E(\vx^j(0)) + r^j(t)$ ~~\# Eq.~\ref{eq:jar}
\STATE $\bar w^j(t) = e^{w^j(t)} / \sum_i e^{w^i(t)}$  \quad \# Self normalized importance sampling
\RETURN $\vx^j(t), \bar w^j(t)$ ~~\# Weighted samples are unbiased estimate of samples from $p(\vx)=e^{-E(\vx)}/Z$
\end{algorithmic}
\caption{Jarzynski sampling}
\label{alg:jar}
\end{algorithm}
\end{minipage}
\end{figure}

\subsection{Model Architecture}\label{sec:architecure}

We use the following architectures from~\cite{nijkamp1} for training EBMs. Convolutional operation $conv(n)$ with $n$ output feature maps and bias term and $fc(n)$ is a fully connected layer. Leaky-ReLU nonlinearity $LReLU$ with default leaky factor $0.05$. We set $n_f = 128$.

\begin{table}[H]
	\small
	\begin{center}
		\def\arraystretch{1.5}
		\setlength{\tabcolsep}{2pt}
		\begin{tabular}{ccc}
			\specialrule{.1em}{.05em}{.05em}
			\multicolumn{3}{c}{Energy-based Model $(32\times 32\times 3)$} \\
			\hline
			Layers & In-Out Size & Stride \\
			\hline
			Input & $32\times32\times3$ & \\
			$3\times3$ conv($n_f$), LReLU & $32\times32\times n_f$ & 1 \\
			$4\times4$ conv($2*n_f$), LReLU & $16\times16\times (2*n_f)$ & 2\\
			$4\times4$ conv($4*n_f$), LReLU & $8\times8\times (4*n_f)$ & 2 \\
			$4\times4$ conv($8*n_f$), LReLU & $4\times4\times (8*n_f)$ & 2 \\
			$4\times4$ conv(1) & $1\times1\times1$ & 1\\
			\specialrule{.1em}{.05em}{.05em}
		\end{tabular}
	\end{center}
	\caption{Network structures $(32\times 32\times 3)$.}
	\label{tab:modelcifar}
\end{table}

\begin{table}[H]
	\small
	\begin{center}
		\def\arraystretch{1.5}
		\setlength{\tabcolsep}{2pt}
		\begin{tabular}{cc}
			\specialrule{.1em}{.05em}{.05em}
			\multicolumn{2}{c}{Energy-based Model $(2,)$} \\
			\hline
			Layers & In-Out Size \\
			\hline
			Input & $2$ \\
			 fc($n_f$), LReLU & $ n_f$  \\
			 fc($2*n_f$), LReLU & $ (2*n_f)$ \\
			 fc($2*n_f$), LReLU & $(2*n_f)$  \\
			 fc($2*n_f$), LReLU & $(2*n_f)$  \\
			 fc(1) & $1$ \\
			\specialrule{.1em}{.05em}{.05em}
		\end{tabular}
	\end{center}
	\caption{Network structure for 2-D toy datasets.}
	\label{tab:model32}
\end{table}

\subsection{Hyper-parameters}\label{sec:hyper}

\paragraph{Table~\ref{tab:ess} and Fig.~\ref{fig:mmd_esh_comp} Experiments} For HMC, we used a step size of $\eps=0.01$ with $k=5$ steps per chain. For MALA and ULA we used both step sizes of $\eps=\{0.01, 0.1\}$. For NUTS we used the default settings from the littleMCMC implementation linked in Sec.~\ref{sec:license}. For the Nose-Hoover sampler we used a step size of $\eps=0.01$ because larger step sizes resulted in numerical errors. For the ESH Leapfrog sampler we used $\eps=0.1$ because we noted from the comparison of ESH ODE solvers that smaller values were less effective, and $\eps=1$ sometimes led to numerical errors.

\paragraph{Sampling pre-trained JEM model} We used the same step size used for training the JEM model for both MALA and ULA, $\eps = 0.01$. 
The Langevin dynamics is defined as,
$$\vx_{t+1} = \vx_{t} + \nicefrac{\eps^2}{2} \vg(\vx_t) + \eps \vv_t,$$
where $\vv_t \sim \mathcal N(0,1)$ at each step. 
However, in JEM and EBM training the energy model and its gradients are implicitly scaled according to the step size, $E(\vx) \rightarrow E(\vx) / (\nicefrac{\eps^2}{2})$ so that the effective update is as follows. 
$$\vx_{t+1} = \vx_{t} + \vg(\vx_t) + \eps \vv_t.$$
For sampling with other methods, we also scaled the energy function as is done for Langevin dynamics. For the ESH leapfrog sampler we used step sizes of $\eps=\{0.5, 1\}$. 

\paragraph{Training EBMs on 2-D Toy Models} We use the architecture in Table~\ref{tab:model32} with $n_f=32$ for all experiments training neural energy models on 2-D datasets. We trained for 10000 iterations with a batch size of 100. We used SGD for optimization with a learning rate of 0.1. We ran each chain for 50 steps. For rings, we initialized from a scaled unit Gaussian with variance 4 and for the mixture of Gaussians we initialized from the uniform distribution in $[-2,2]$. As suggested by \cite{nijkamp2}, we used ULA for sampling with a step size of $\eps=0.1$. For ESH we also used a step size $\eps=0.1$.

\paragraph{CIFAR EBM details}
For the experiment training an EBM on CIFAR data in Sec.~\ref{sec:cifar} we chose to match hyper-parameters as closely as possible to EBM training with Langevin dynamics in a regime where it is known to be stable. 
We use the architecture in Table~\ref{tab:modelcifar} with $n_f=128$. 
We add Gaussian noise with variance $0.03$ as done by \cite{nijkamp1} to stabilize training. 
We use chains with $100$ steps each. 
We used a replay buffer of size $10,000$. 
Following \cite{du} we applied spectral normalization~\cite{sngan}, and we also ensemble by averaging the energy over the last 10 epoch checkpoints. 
The optimizer was ADAM with a learning rate of $0.0002$ with $\beta_1=0.5, \beta_2=0.999$.  
The buffer was initialized uniformly over $[-1,1]$. 
We used the same scaled energy function for both Langevin and ESH sampler. 
The batch size was $500$ and we trained for $250$ epochs. 
ULA step size was $\eps=0.01$ and ESH step size was $\eps=0.5$.

\paragraph{Test time sampling}
In Fig.~\ref{fig:cifar} we show samples generated in exactly the same manner as training, except with a fixed energy function. 
For sampling from scratch, we initialize the sampler with an equal mixture of a random constant color and per pixel standard normal noise. The random constant color is drawn from a normal distribution constructed from the pixel-averaged image color of CIFAR training images. We also clipped color values to be in the range $[-1,1]$. 

\paragraph{Jarzynski Sampler EBM Training in Sec.~\ref{sec:jarebm}} 
We use the architecture in Table~\ref{tab:modelcifar} with $n_f=128$ for training a neural energy model on CIFAR data in Sec.~\ref{sec:train}. We did not add noise to the data as was done by \cite{nijkamp1} to stabilize training, to highlight that this step is not necessary for unbiased Jarzynski samplers. We use short chains with $50$ steps each. We used a replay buffer with the same size as the training data, $50,000$. We trained for 30 epochs over the training data with batch sizes of 1000. The optimizer was ADAM with a learning rate of $10^{-3}$ and $\beta_1 =0.9, \beta_2=0.999$.  Chains were initialized from a persistent buffer that was initialized uniformly over $[-1,1]$. 
Note that when we initialize ESH dynamics with a sample from the buffer an ambiguity arises in setting the value of $E_0(\vx(0))$ in Eq.~\ref{eq:jar}. We considered $E_0$ to be constant. 
For the Langevin sampler we scaled the energy function by $2/\eps^2$, but for ESH sampling we skipped this step to show it is not necessary. 
ULA sampling used a step size of $\eps=0.01$ (with larger gradient steps due to energy scaling), and ESH used a step size $\eps=0.5$. Although a step size for ESH of $\eps=1$ worked better for sampling the pre-trained JEM model, the error from discretizing the ODE goes like $O(\eps^3)$ so we erred on the conservative side by using $\eps < 1$. When we tried $\eps >1$ we always got numerical errors, so the apparent stability of the choice $\eps=1$ may require more consideration.

\subsection{Computational Cost}\label{sec:cost}
Our small experiments comparing samplers and ablation study were done on a single NVIDIA 1080Ti GPU and took from minutes to hours. The Jarzynski sampler trained CIFAR models were trained using 4 Tesla V100 GPUs for about 3 hours for each experiment.  The CIFAR EBM in Sec.~\ref{sec:cifar} trained using 4 Tesla V100 GPUs for about 1 day for each experiment.

\subsection{Code Used and Licenses}\label{sec:license}
Our code is provided at \url{https://github.com/gregversteeg/esh_dynamics}. We implemented our approach in PyTorch and also used code from a number of sources. 

\begin{tabular}{l c c}
\underline{Link} & \underline{Citation} & \underline{License} \\
 \url{https://github.com/point0bar1/ebm-anatomy} & \cite{nijkamp1} &   MIT License \\
   \url{https://github.com/rtqichen/torchdiffeq} & \cite{chen2018neural} & MIT License \\
  \url{https://github.com/wgrathwohl/JEM} & \cite{secret_classifier}& Apache version 2 \\
  \url{https://github.com/eigenfoo/littlemcmc}& \cite{nuts} & Apache version 2 
\end{tabular}

\section{Additional Results}\label{app:cifar}

\subsection{Comparing Integrators for ESH}\label{app:integrate}

\begin{figure}[htbp]
    \centering
    \includegraphics[width=0.5\textwidth]{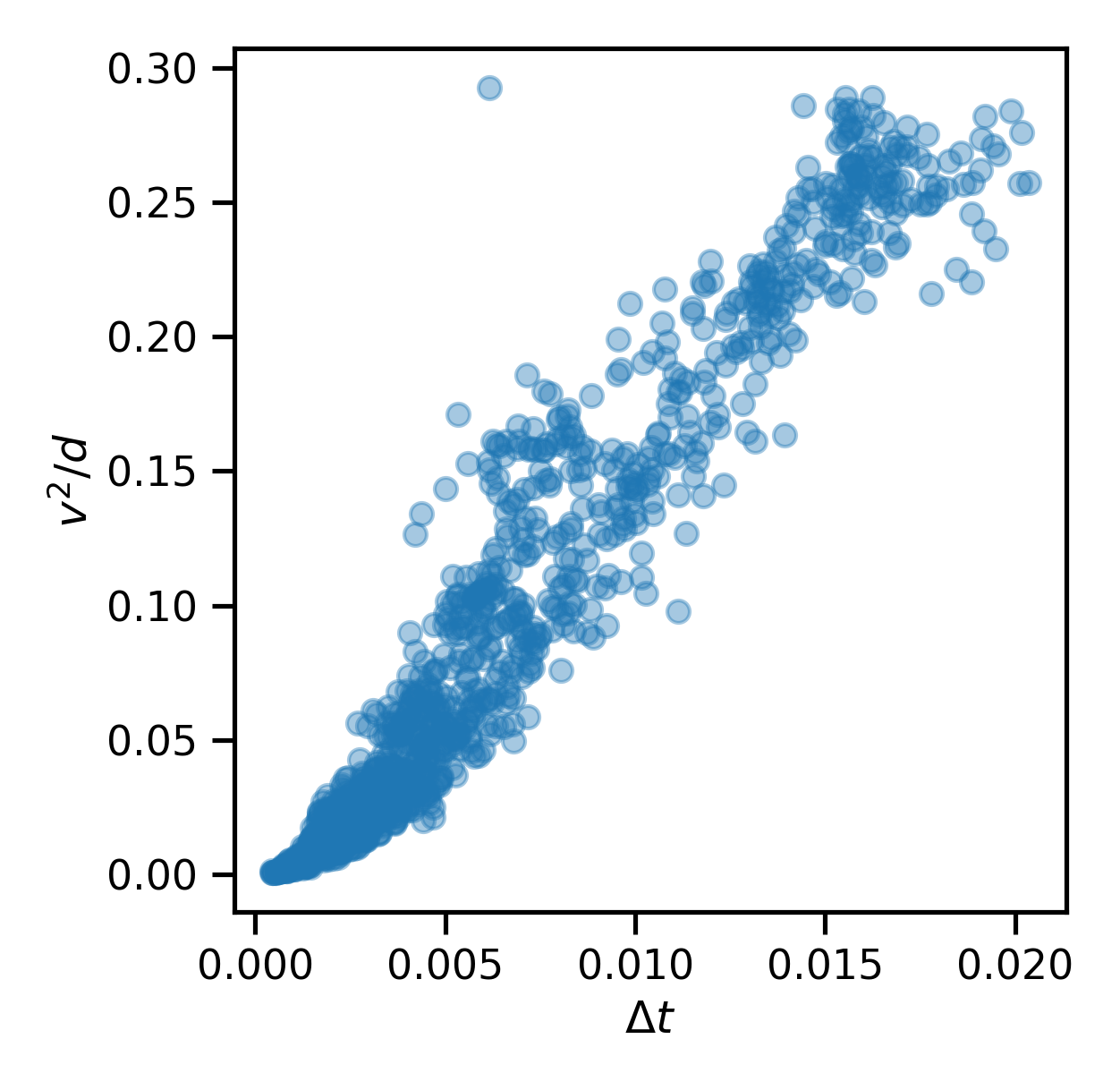}
    \caption{In the original unscaled ODE, Eq.~\ref{eq:dynamics}, the adaptive time-step from the Runge-Kutta integrator is correlated with the magnitude of the velocity. This motivates our idea to re-scale the time-step.}
    \label{fig:scale_t}
\end{figure}

\begin{figure}[htbp]
    \centering
    \includegraphics[width=\textwidth]{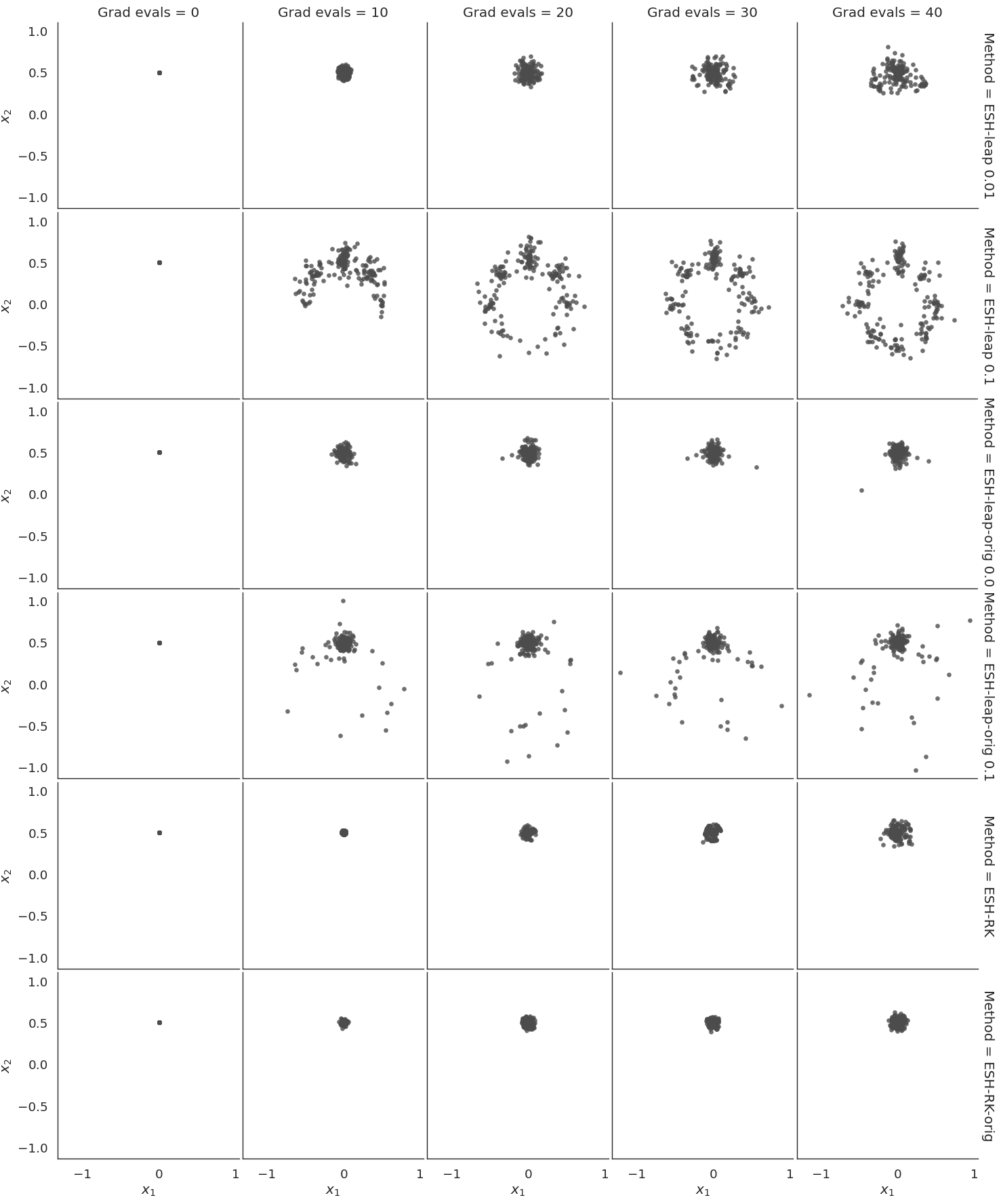}
    \caption{For the 2d mixture of Gaussians with an informative prior, we visualize the distribution of 500 chains after various numbers of gradient evaluations, comparing different ESH dynamic integrators.}
    \label{fig:esh_movie}
\end{figure}

\begin{figure}[htbp]
    \centering
        \begin{tabular}{c c c}
        \tiny \textbf{2D MOG} & \tiny \textbf{2D MOG-Prior} & \tiny \textbf{2D SCG} \\
    \includegraphics[width=0.3\textwidth,trim={0 0 7cm 0},clip]{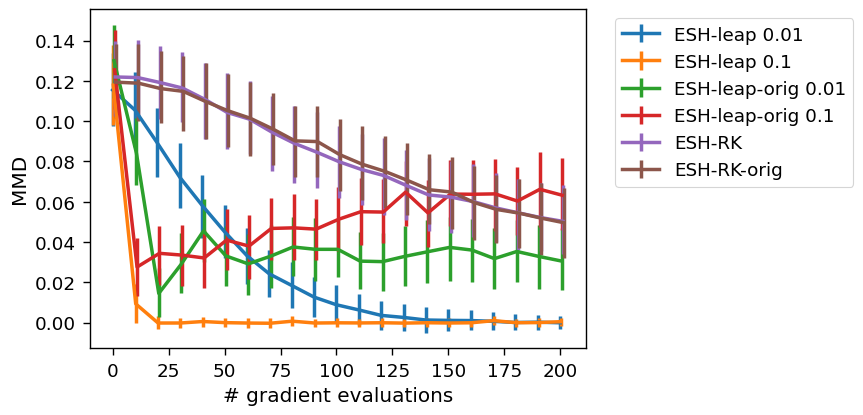} &
    \includegraphics[width=0.3\textwidth,trim={0 0 7cm 0},clip]{figures/ablation/mmd_2D-MOG-prior.png} &
    \includegraphics[width=0.3\textwidth,trim={0 0 7cm 0},clip]{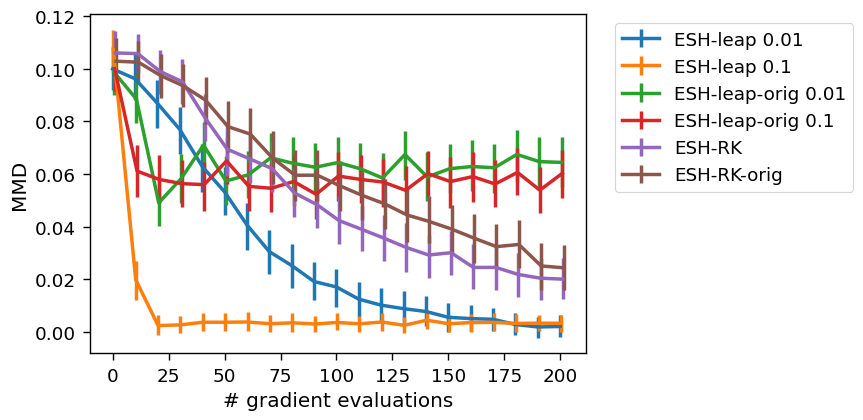} \\
          \tiny \textbf{2D SCG-Bias} & \tiny \textbf{50D ICG} &  \\
    \includegraphics[width=0.3\textwidth,trim={0 0 7cm 0},clip]{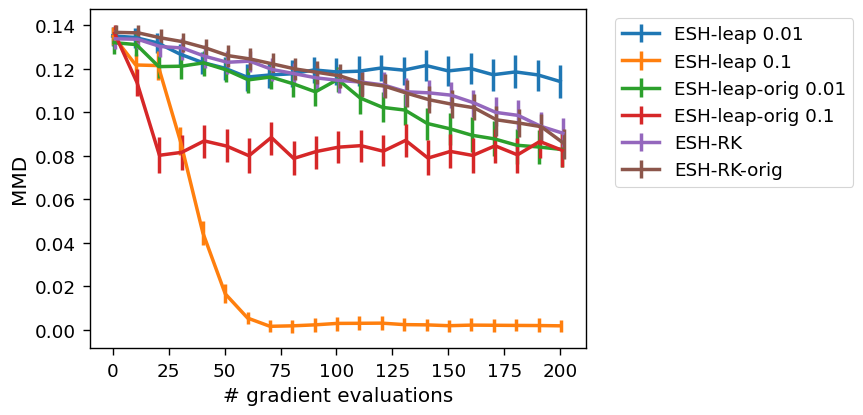} &
    \includegraphics[width=0.3\textwidth,trim={0 0 7cm 0},clip]{figures/ablation/mmd_50D-ICG.png} &
    \includegraphics[width=0.25\textwidth,trim={15.5cm 5cm 0 0},clip]{figures/ablation/mmd_50D-ICG.png}
     \end{tabular}
    \caption{Maximum Mean Discrepancy (MMD) as a function of the number of gradient evaluations for other datasets using different ESH integrators.}
    \label{fig:mmd_esh_comp}
\end{figure}

We provide additional figures about how different ESH dynamics integrators perform. In Fig.~\ref{fig:scale_t} we show how the adaptive step size in the Runge-Kutta integrator correlates with velocity, motivating our attempt to scale the ESH dynamics. Fig.~\ref{fig:esh_movie} shows the evolution of a batch of 500 chains evolving using different integrators. You can see that the ESH leapfrog with scaled dynamics explores the space much more rapidly than the other methods. 
In Fig.~\ref{fig:mmd_esh_comp} we show the MMD scores for various integrators on our test sets. The scaled leapfrog integrator dominates across all experiments.

\begin{figure}[htbp]
    \centering
    \includegraphics[width=\textwidth]{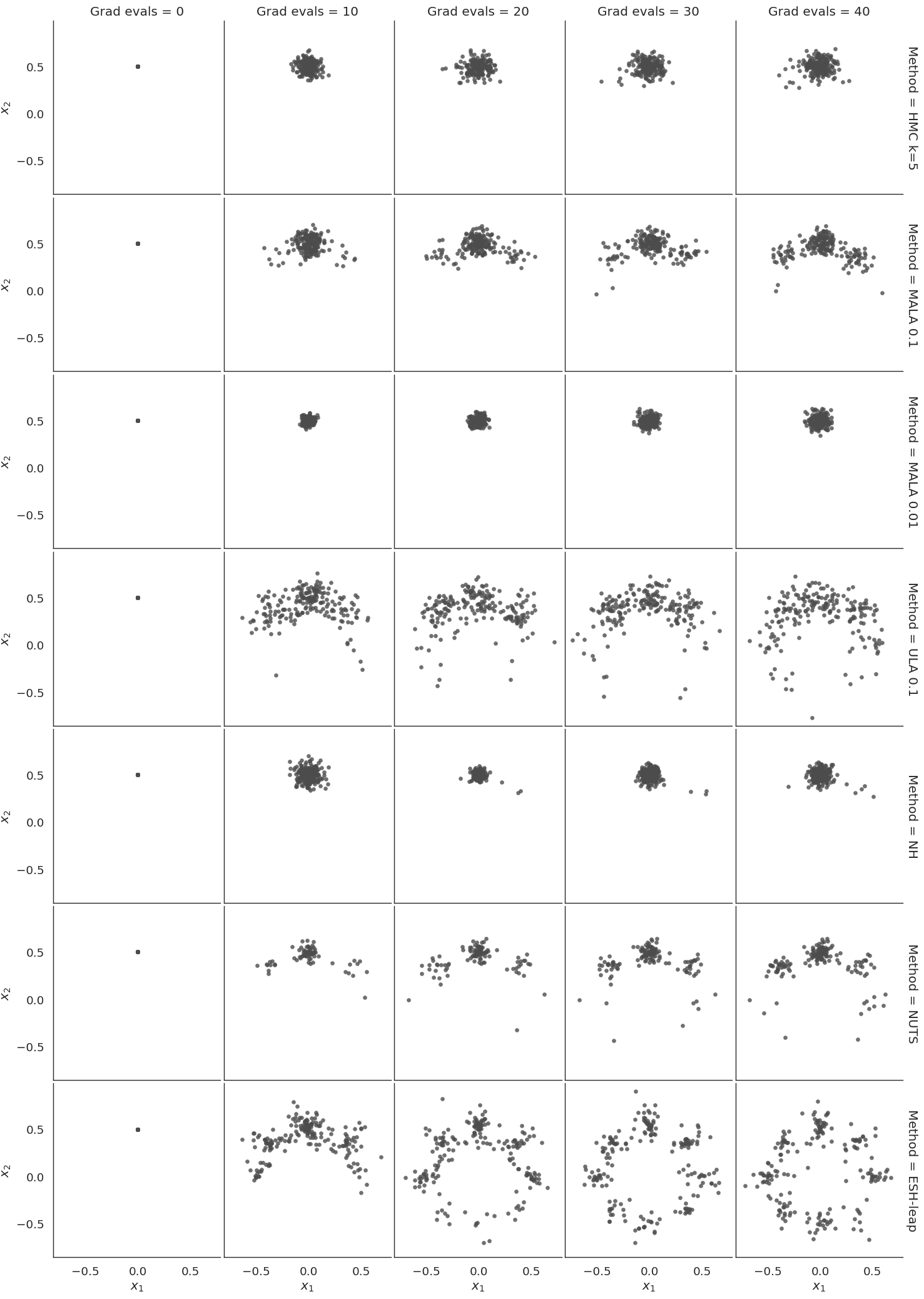}
    \caption{For the 2d mixture of Gaussians with an informative prior, we visualize the distribution of 500 chains after various numbers of gradient evaluations, comparing different samplers.}
    \label{fig:esh_movie2}
\end{figure}

\begin{figure}[htbp]
    \centering
    \begin{tabular}{l c c c}
            & \tiny \textbf{ESH Leap (scaled ODE)} & \tiny \textbf{ESH RK(scaled ODE)} & \tiny \textbf{ESH RK (original ODE)} \\
    \raisebox{1.5cm}{\tiny $H(\vx,\vv)$} &\includegraphics[width=0.3\textwidth]{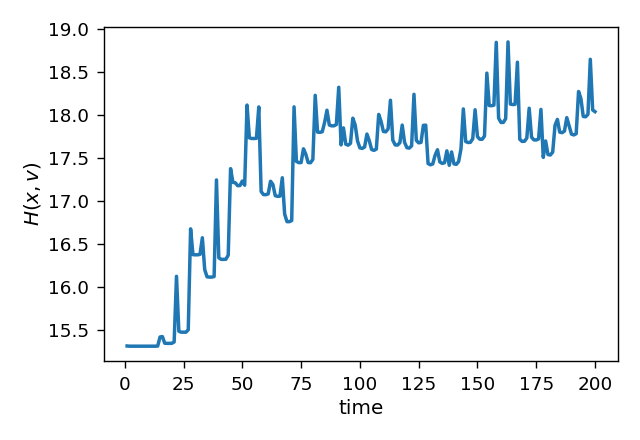} &
        \includegraphics[width=0.3\textwidth]{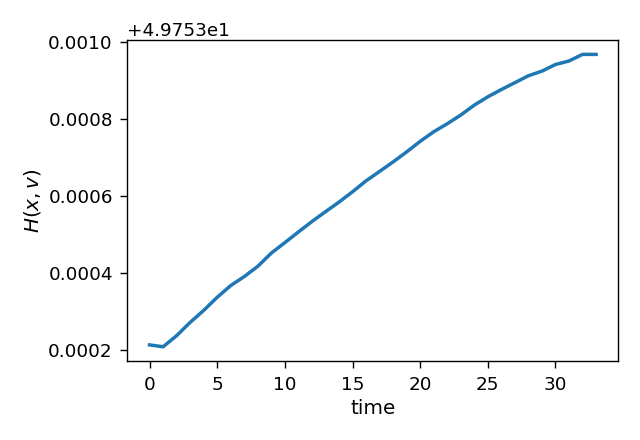} &
                    \includegraphics[width=0.3\textwidth]{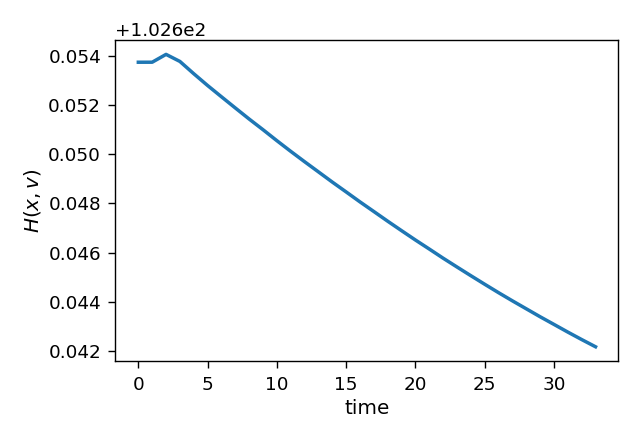} \\
    \raisebox{1.5cm}{\tiny $K(\vv)$} &        \includegraphics[width=0.3\textwidth]{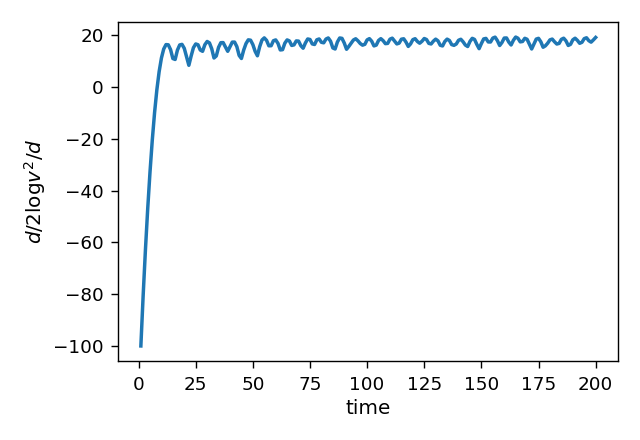}  &
                \includegraphics[width=0.3\textwidth]{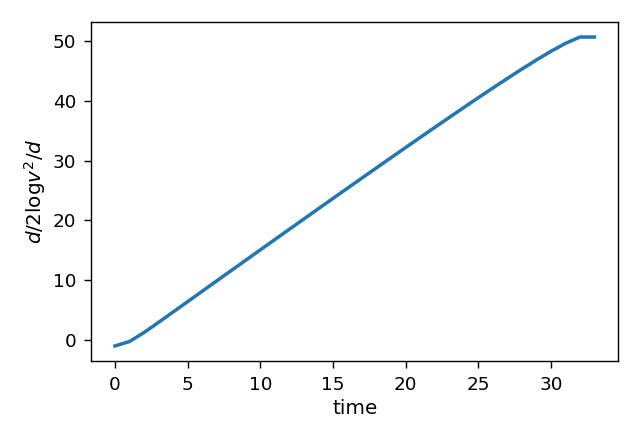} &
        \includegraphics[width=0.3\textwidth]{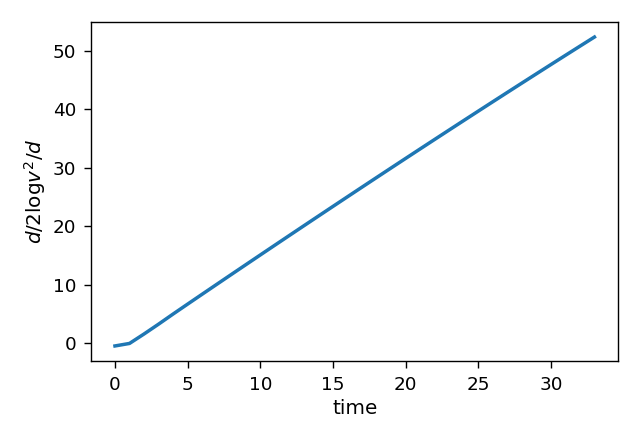} 
    \end{tabular}
    \caption{For the MOG dataset, we visualize the total Hamiltonian and the kinetic energy for different ESH integrators. Note the small relative scale on the $y$-axis for the total Hamiltonian. The step size is $\eps=0.1$.}
    \label{fig:ham}
\end{figure}

In Fig.\ref{fig:ham} we show the total Hamiltonian and kinetic energy for different ESH integrators. They all approximately conserve the Hamiltonian, but only the leapfrog integrator approaches a low energy, high kinetic energy solution quickly. Quickly approaching this regime followed by small oscillations in kinetic energy is a typical characteristic of the dynamics.

\subsection{Visualizing Ergodicity}\label{sec:ergodic}

\begin{figure}[htbp]
    \centering
        \includegraphics[width=0.5\textwidth]{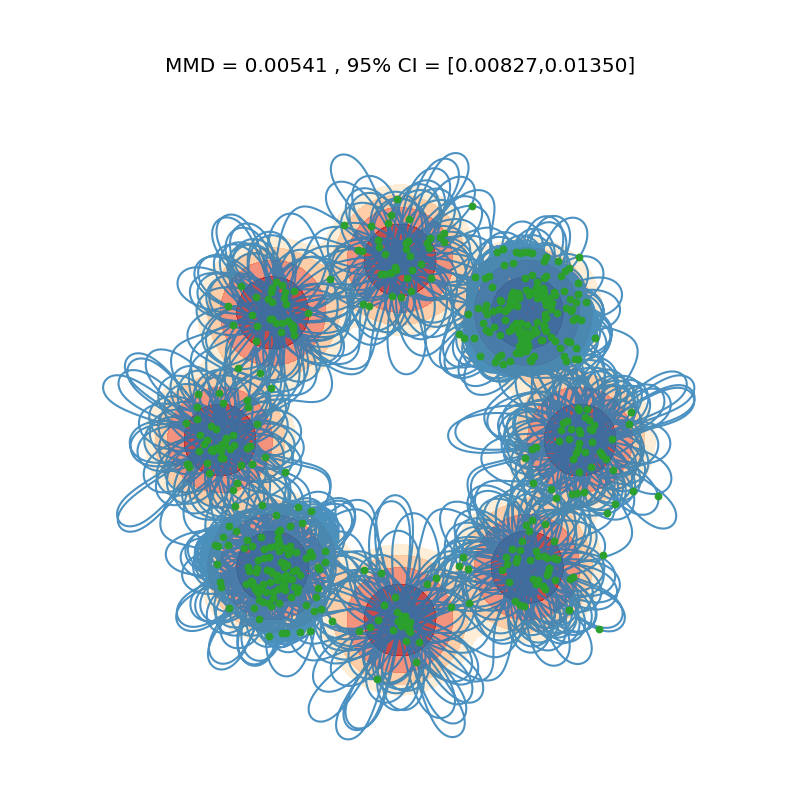} 
    \caption{For the 2D MOG dataset, we use ESH-Leap to simulate a long trajectory ($500,000$ steps) with small step size ($\eps=0.001$) for a precise simulation. We ergodically sample 500 points from this single trajectory and compute the MMD to the ground truth. }
    \label{fig:ergodic}
\end{figure}

To empirically visualize whether the ESH dynamics are ergodic, we simulated a single long trajectory on one of our 2D datasets, MOG. We used a small step size, $\eps=0.001$, to accurately simulate the dynamics. It is possible that larger step sizes introduce errors in the trajectory, but that these errors actually help the dynamics mix. The reason is that the dynamics are supposed to sample the target distribution for any energy from any initial point. Therefore, discretization error may actually help by mixing between different trajectories, any of which ergodically sample the target distribution.  We want to be sure that the mixing comes from the dynamics, and not discretization error, for this test. Then, to quantify how well the dynamics converge to the correct distribution, we ergodically sample 500 points and estimate the MMD to samples from the ground truth distribution as in Sec.~\ref{sec:results}. As shown in Fig.~\ref{fig:ergodic}, the trajectory seems to be ergodic, i.e. it visits all points in the space and does not get stuck in a subspace. This is quantified by the MMD which is close to zero as expected. The trajectory does not appear to have any regularity or cycles. 
Note that ergodicity does not necessarily imply chaotic dynamics, which require that small changes in initial conditions lead to strongly diverging trajectories as measured by Lyapunov exponents, for instance. We did not test how chaotic the ESH dynamics are, but this would be interesting to study in future work, especially as it relates to mixing speed for sampling.  

\subsection{Funnel}\label{app:funnel}
\begin{figure}[htbp]
    \centering
        \includegraphics[width=0.59\textwidth]{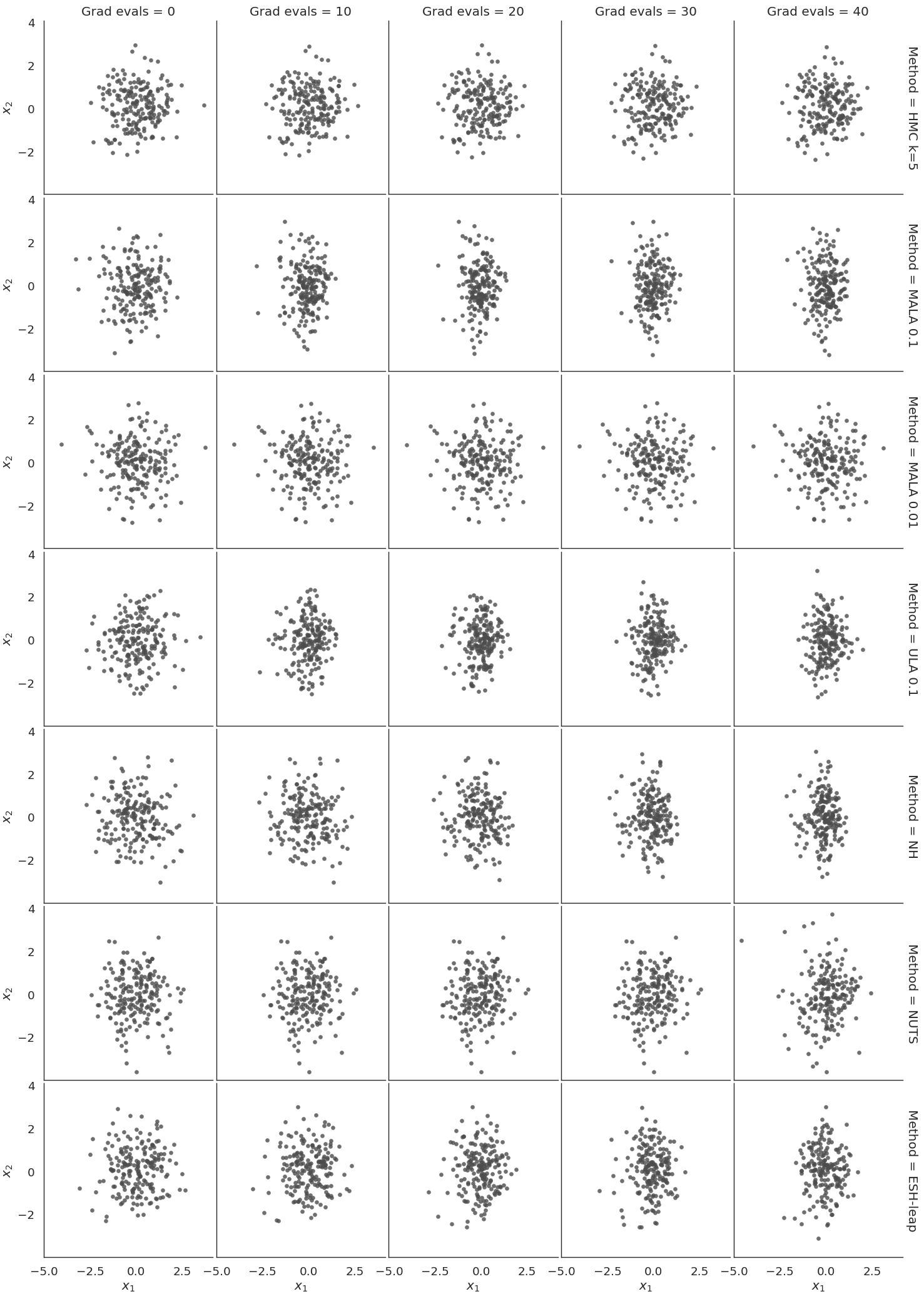}     \includegraphics[width=0.4\textwidth,trim={0 0 0 0},clip]{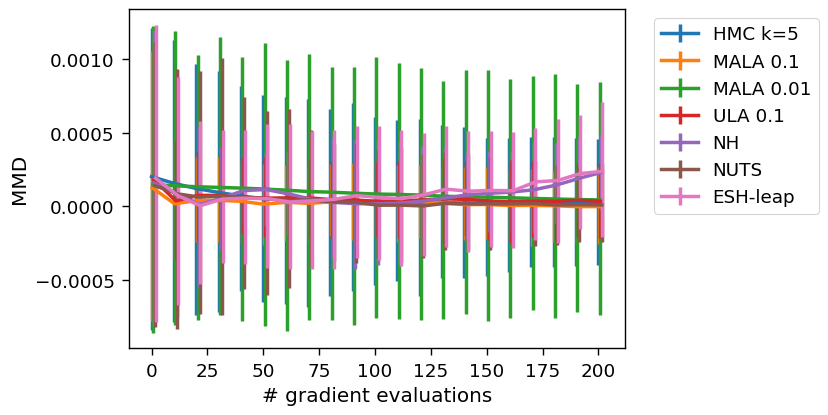}
    \caption{For the funnel dataset, $x_1$ ($x$-axis) controls the width of the funnel. We should see a thin funnel toward the left of the plot. We visualize the distribution of 500 chains after various numbers of gradient evaluations.}
    \label{fig:funnel}
\end{figure}

All methods failed on the Funnel dataset. The MMD plots in Fig.~\ref{fig:funnel} are ambiguous, but from looking at the evolution we see that none of the samplers have captured the funnel shape. For a dataset where the length scales differ by orders of magnitude, auxiliary models or second order methods may be justified. 

\subsection{Sampling a Pre-trained JEM Model with Informative Prior}
\begin{figure}[hbp]
    \centering
        \begin{minipage}{.7\textwidth}
        \includegraphics[width=0.99\textwidth]{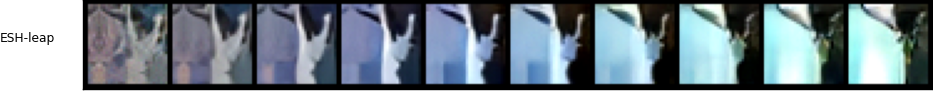}
            \includegraphics[width=0.99\textwidth]{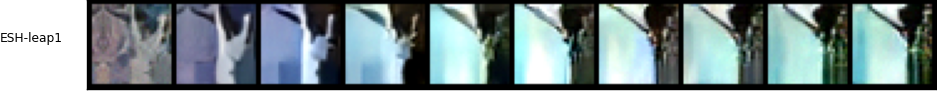}
                        \includegraphics[width=0.99\textwidth]{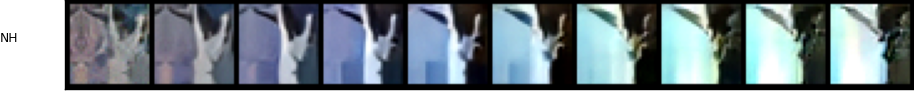}
                                                \includegraphics[width=0.99\textwidth]{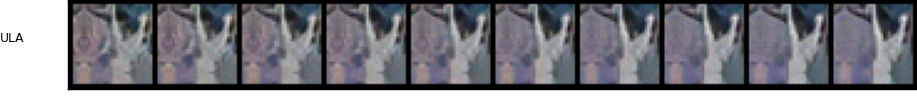}
    \end{minipage} 
    
    \includegraphics[width=0.7\textwidth]{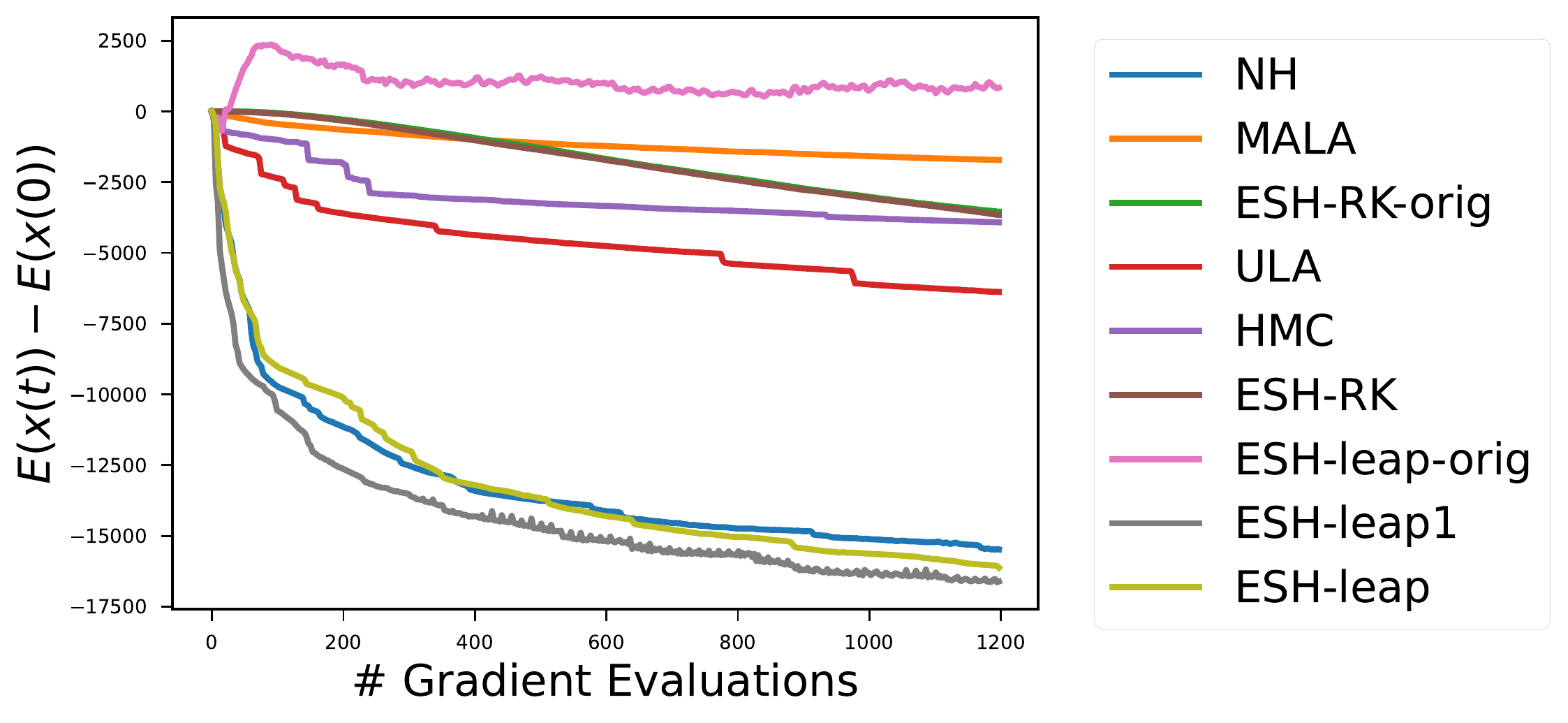}
    \caption{(Top) Example of sampling chains from replay buffer initialization with 200 gradient evaluations per method. (Bottom) Average energy over time for a batch of 50 samples using different samplers.}
    \label{fig:prior_jem}
\end{figure}

In Fig.~\ref{fig:jem}, we showed that ESH samplers found much lower energy solutions than other samplers when initialized from noise with a pre-trained neural network energy model. 
However, JEM was trained with persistent contrastive divergence, so it makes sense to compare the solutions found when starting from a sample in the persistent buffer. Fig.~\ref{fig:prior_jem} shows that the result is the same, ESH finds much lower energy solutions. 

\end{document}